\documentclass[10pt,twocolumn,letterpaper]{article}

\usepackage[pagenumbers]{wacv} 

\definecolor{wacvblue}{rgb}{0.21,0.49,0.74}
\usepackage[pagebackref,breaklinks,colorlinks,allcolors=wacvblue]{hyperref}
\usepackage{multirow}
\usepackage[most]{tcolorbox} 
\usepackage{tabularray}
\usepackage{siunitx}

\newtcolorbox{inlinenote}[2][]{%
  enhanced,
  breakable,
  colback=#2!10,
  colframe=#2!60!black,
  left=1pt,right=1pt,top=1pt,bottom=1pt,
  boxrule=0.6pt,arc=2pt,
}

\usepackage[table]{xcolor}
\definecolor{best}{rgb}{0.70, 0.88, 0.70}   
\definecolor{second}{rgb}{0.85, 0.95, 0.85} 

\usepackage{soul}
\soulregister\cite7
\soulregister\ref7
\soulregister\pageref7
\soulregister\cref7
\soulregister\Cref7

\definecolor{addedgreen}{HTML}{D9F2D9}
\definecolor{changedyellow}{HTML}{FFF2B2}

\sethlcolor{addedgreen}

\def\wacvPaperID{56} 
\def\confName{WACV}
\def\confYear{2027}

\title{GraphPilot: Grounded Scene Graph Conditioning for Language-Based Autonomous Driving}

\author{
Fabian Schmidt\textsuperscript{1,2} \qquad
Markus Enzweiler\textsuperscript{1} \qquad
Abhinav Valada\textsuperscript{2}\\[10pt]
\textsuperscript{1}Esslingen University of Applied Sciences \qquad
\textsuperscript{2}University of Freiburg\\
}

\begin{document}
\maketitle

\begin{abstract}
Vision-language models have recently emerged as promising planners for autonomous driving, where success hinges on reasoning over spatial structure and dynamic interactions from multimodal input.
However, existing models are typically trained without supervision that explicitly encodes relational dependencies, requiring them to infer interactions between traffic entities implicitly from raw sensor data.
In this work, we introduce GraphPilot, a model-agnostic approach that conditions language-based driving models on structured relational context in the form of traffic scene graphs.
We serialize scene graphs at different abstraction levels and formats and inject them through structured prompts without modifying the underlying architecture or training objective.
Extensive closed-loop evaluations on LangAuto and Bench2Drive show persistent improvements across LMDrive, BEVDriver, and SimLingo.
Controlled ablations and hidden-state probing further indicate that relational supervision contributes substantially to these improvements while increasing the decodability of scene relations.
Open-loop SpaceDrive results on nuScenes further demonstrate transfer to real-world data.
These results indicate that diverse architectures can effectively internalize and ground relational priors through scene graph-conditioned training, even without requiring scene graph input at test-time.
Code, fine-tuned models, and our scene graph dataset are publicly available at \url{https://github.com/iis-esslingen/GraphPilot}.
\end{abstract}

\section{Introduction}
\label{sec:intro}

\begin{figure}[t]
  \centering
   \includegraphics[width=1.\linewidth]{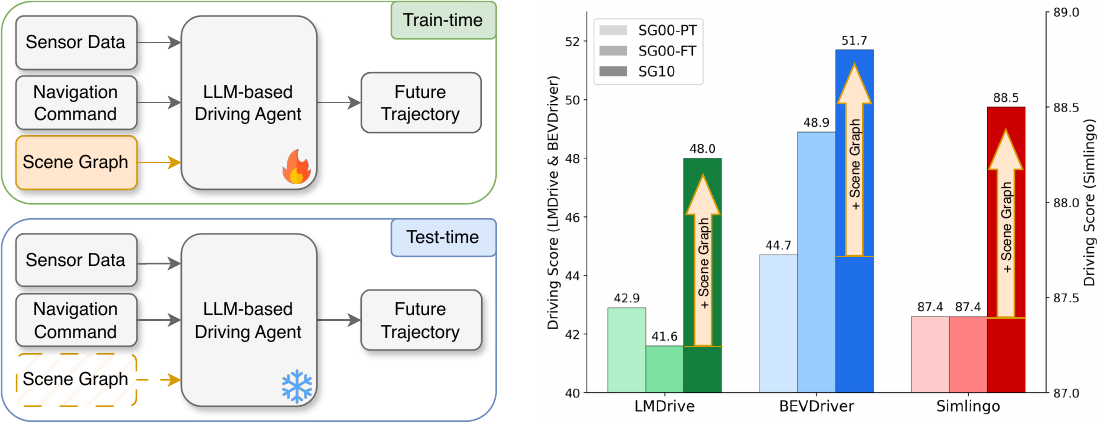}
   \caption{\textbf{Explicit relational grounding through scene graph conditioning.} We visualize three setups: SG00-PT (the original pretrained baseline), SG00-FT (the baseline fine-tuned without scene graphs), and SG10 (conditioned on scene graphs during training but absent at test-time). The substantial performance gains across LMDrive, BEVDriver, and SimLingo suggest that models internalize relational structure during training, even when scene graphs are absent at test-time.
   \looseness=-1}
   \label{fig:overview}
   \vspace{-1.0em}
\end{figure}

Understanding complex traffic scenes remains a central challenge in autonomous driving, especially when decision-making is conditioned on natural language instructions~\cite{cui2024survey}. 
Modern language-based driving agents operate at the intersection of perception, language, and planning, requiring not only accurate scene understanding but also the ability to reason about spatial structure, traffic rules, and interactions among dynamic actors~\cite{xie2025vlms-ready}. 
While recent works have shown that vision-language models (VLMs) can map image and language inputs to plausible trajectories~\cite{shao2024lmdrive,winter2025bevdriver,renz2025simlingo, zhang2025vldrive, zhang2025adadrive, schmidt2025enhancing, schmidt2026laddrive}, their performance in complex environments remains limited. 
A key limitation is that current models typically rely on representations that do not explicitly encode relational structure, forcing them to infer safety-critical interactions implicitly from dense features~\cite{zhang2025safeauto}.

Scene graphs provide an expressive solution to this limitation~\cite{johnson2015image,kassab2024bare}. 
By explicitly representing a traffic scene as a structured graph of entities and their relations, they encode not only \textit{what} is present in the environment, but also \textit{how} entities are related and may influence one another. 
This relational representation enhances semantic understanding and enables more informed planning~\cite{chen2025static} by mapping physical entities to nodes and expressing their spatial, semantic, or regulatory dependencies as edges.

Despite their growing use in indoor navigation and instruction-following tasks~\cite{gu2024conceptgraphs,rana2023sayplan,yin2024sgnav,honerkamp2024language,werby2024hierarchical,mohammadi2025more}, scene graphs have only recently gained attention in outdoor autonomous driving, mainly for tasks such as risk estimation~\cite{malawade2022sg2vec,liu2023learning}, and trajectory prediction~\cite{jia2023hdgt,zhang2025graphad}. However, their integration into language-based planning remains unexplored. Instead, most existing systems represent the scene using dense geometric or visual features, including image tokens~\cite{renz2025simlingo, zhang2025vldrive, zhang2025adadrive} or BEV grids~\cite{shao2024lmdrive, winter2025bevdriver, hegde2025dima}, followed by generic attention mechanisms. 
As a result, critical relationships between entities must be inferred implicitly. 
This gap motivates our central question: Can \textit{explicit} relational structure of a traffic scene help language-based driving models internalize and reason over spatial and relational dependencies in traffic scenes?

To address this question, we introduce \textbf{GraphPilot}, a model-agnostic approach for injecting structured semantic context into language-based autonomous driving systems. We construct traffic scene graphs at each time step and serialize them into human-readable formats (Text, JSON, YAML), which are then provided to the language model along with the navigation instruction and sensory input, as shown in \cref{fig:overview}. We make this a deliberate design choice to avoid changes to the model architecture, additional training objectives, or specialized graph encoders, thereby maintaining model-agnostic applicability and ease of integration. To the best of our knowledge, we present the first work to inject a traffic scene graph as explicit context for planning in a language-based driving model. 


Our experiments reveal several key insights. 
First, using scene graph prompts at test-time alone improves performance over the original models. 
Second, training with scene graph supervision yields substantial improvement in driving performance, and importantly, these improvements persist even when scene graphs are omitted at test-time. 
These findings suggest that language-based planners can internalize relational knowledge during training and retain its benefits under graph-free inference.
Controlled ablations and representation probing further show that relational content contributes substantially to these improvements and becomes more decodable in the resulting hidden states.
Finally, we find that leaner scene graph abstractions offer a strong trade-off between semantic fidelity and prompt efficiency, achieving high performance with fewer tokens.


Our \textbf{contributions} are threefold:
(i) \textbf{Model-agnostic relational grounding}, introducing structured traffic scene graphs into vision-language planners without architectural changes, specialized graph encoders, or auxiliary objectives.
(ii) \textbf{Relational internalization}, enabling persistent graph-free performance gains through training-time scene graph supervision.
(iii) \textbf{Efficient scene representation}, systematically analyzing abstraction levels, serialization formats, and prompt designs to identify compact and effective representations for planning.
We validate our approach across LMDrive, BEVDriver, SimLingo, and SpaceDrive on closed-loop CARLA benchmarks and open-loop evaluation on nuScenes.

\section{Related Work}
\label{sec:related_work}

{
\noindent\textbf{Language-Based Autonomous Driving}: 
Several works deploy VLMs as planners that map sensor data and navigation commands directly to future waypoints~\cite{shao2024lmdrive,winter2025bevdriver,renz2025simlingo,zhang2025vldrive,xu2025drivegpt4v2,xie2025s4driver,li2026spacedrive}, spanning both open-loop trajectory prediction on nuScenes~\cite{caesar2020nuscenes} and closed-loop evaluation via CARLA~\cite{dosovitskiy2017carla} or NavSim~\cite{dauner2024navsim}.
LMDrive~\cite{shao2024lmdrive} presents a VLM-based end-to-end stack that fuses camera and LiDAR inputs with natural-language navigation commands, while BEVDriver~\cite{winter2025bevdriver} extends this design by strengthening spatial grounding through richer BEV features.
More recently, SpaceDrive~\cite{li2026spacedrive} introduces unified 3D positional encodings for visual and coordinate representations to improve spatial reasoning.
Beyond direct waypoint prediction, some approaches aim to increase reasoning fidelity by incorporating chain-of-thought within a perception-prediction-planning paradigm, accepting additional latency in exchange for interpretability and more transparent decision making~\cite{kong2025vlrdriver,zeng2025futuresightdrive,hwang2025emma,tian2024drivevlm}. 
Efficiency-oriented systems reduce online VLM usage: AdaDrive~\cite{zhang2025adadrive} employs an adaptive slow-fast collaboration that determines when the VLM should be involved.
Other approaches decouple high-level reasoning from trajectory generation, using VLM-generated subgoals, constraints, or safety signals to guide non-VLM planners~\cite{fu2025orion,chen2025solve,han2025dme,pan2024vlp,xu2024vlmad}. 
Finally, some methods transfer VLM knowledge into training-time supervision to avoid online VLM inference~\cite{hegde2025dima,zhang2024feedback}, while RAG-based approaches retrieve similar scenarios or past experiences to improve robustness on rare events~\cite{yuan2024ragdriver,mei2024leapad,wen2024dilu,wang2025rad,zhang2025safeauto}.
}

\begin{figure*}
    \centering
    \includegraphics[width=\textwidth]{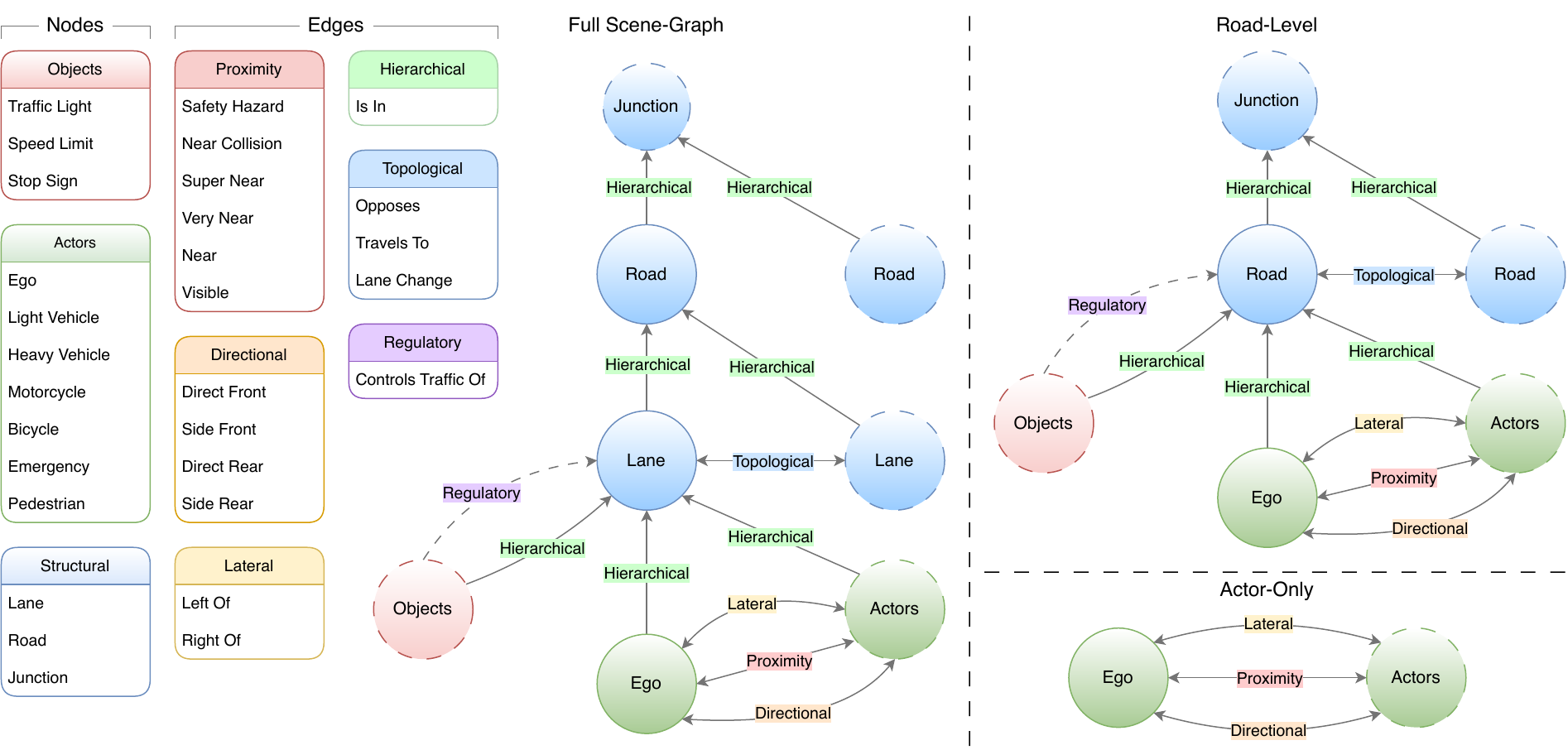}
    \caption{\textbf{Scene graph construction.} Each traffic scene is represented as a structured, labeled graph capturing entities (actors, objects, structure) and their relations. We define three levels of abstraction: \textit{Full} (all node types and relations), \textit{Road-Level} (collapsed structural detail), and \textit{Actor-Only} (actors and their pairwise interactions), enabling analysis of the trade-off between relational fidelity and prompt efficiency. Dashed nodes and edges indicate optional elements that are not always present.}
    \label{fig:scene_graph_construction}
    \vspace{-0.5em}
\end{figure*}

{
\noindent\textbf{Scene Graphs in Autonomous Driving}:
Traffic scene graphs represent road users, roadway structure, and their relations, with applications spanning graph construction, safety reasoning, retrieval, and trajectory prediction.
Recent work constructs scene graphs from video, multi-camera perception, and HD maps, ranging from rule-based extractors to learned relation predictors, lane-topology models, and multi-agent 3D scene graphs~\cite{malawade2022roadscene2vec,wang2024rs2g,tian2023rsg,lv2025t2sg,tian2025query,greve2024curbsg,steinke2025curbosg,zhang2025parking}.
This progression reflects a shift from hand-crafted relational priors toward learned representations that better capture diverse traffic layouts and interactions.
For safety reasoning, multi-relational GNNs and sequence models capture spatio-temporal interactions for collision prediction and pedestrian risk assessment~\cite{malawade2022roadscene2vec,malawade2022sg2vec,liu2023learning,zhou2024hktsg}.
Retrieval-based methods use subgraph matching or VLM-generated descriptions to identify similar scenes, mine failures, or provide mitigation knowledge through RAG-style augmentation~\cite{tian2023rsgsearch,tian2024rsgsearchplus,tian2025query,liao2025workzones}.
Graph-based trajectory prediction models agent--map interactions with graph or transformer encoders to capture relational dependencies for motion forecasting~\cite{mo2022multi,zhou2022hivt,jia2023hdgt,zhang2025graphad}.
Unlike prior work, we are the first to serialize scene graphs as relational context for language-based planners.
}

\section{Method}
\label{sec:method}

\subsection{Scene Graph Construction}
We first construct a scene graph for each time step by detecting nodes and extracting relations, building on~\cite{woodlief2025closing}.

\noindent \textbf{Formalization.}
We represent the scene at time $t$ as a directed, labeled multigraph $G_t = (V_t, E_t, \mathcal{R})$, where $V_t$ is the set of nodes, $E_t \subseteq V_t \times \mathcal{R} \times V_t$ the set of labeled edges, and $\mathcal{R}$ the relation vocabulary, cf. \cref{fig:scene_graph_construction}.

Nodes carry a unique identifier and a semantic class, partitioned into three disjoint groups such that $V_t = \mathcal{S}_t \cup \mathcal{A}_t \cup \mathcal{O}_t$. 
Structural nodes $\mathcal{S}_t$ encode the static roadway scaffold, including \textit{lanes}, \textit{roads}, and \textit{junctions}, which provide the spatial context and connectivity for motion. 
Actor nodes $\mathcal{A}_t$ represent road users, including the omnipresent \textit{ego} vehicle, dynamic vehicles, and pedestrians, whose interactions define the driving dynamics. 
Traffic objects $\mathcal{O}_t$ capture regulatory context, such as \textit{traffic lights}, \textit{speed limits}, and \textit{stop signs}, which constrain admissible behavior and govern planning decisions.
The relation vocabulary $\mathcal{R}$ is partitioned into six semantic groups that characterize the interactions within the environment. 
Proximity relations quantify distance- and risk-based interactions for collision avoidance, while directional and lateral relations describe longitudinal ordering and side-by-side context to guide lane-keeping or braking decisions. 
The hierarchical relation (\textit{is in}) anchors actors and traffic objects within their structural context to localize local traffic rules. 
Finally, topological relations define lane network connectivity, and regulatory relations link traffic entities directly to the structural elements they govern.
To maintain a compact graph, we impose typing constraints on edge instantiation. The relation \textit{is in} establishes a strict spatial hierarchy (actor $\rightarrow$ lane $\rightarrow$ road $\rightarrow$ junction). Inter-actor edges are limited to behaviorally relevant proximity, directional, and lateral relations, while regulatory edges connect \textit{traffic lights} to their respective lanes via \textit{controls traffic of}.

\noindent \textbf{Abstraction Levels.}
Alongside the full graph, we introduce two abstractions to analyze the trade-off between semantic fidelity and prompt efficiency. \textit{Road-Level} collapses fine-grained lane detail into parent roads by aggregating lane-level memberships and topological connectivity while omitting lane-specific relations like \textit{lane change}. Conversely, \textit{Actor-Only} eliminates all structural and object nodes entirely, restricting the context to local inter-actor geometry by retaining only the ego vehicle, dynamic actors, and their pairwise spatial and directional relations.

\begin{figure*}[t]
  \centering
  \begin{minipage}[b]{0.36\linewidth}
    \centering
    \includegraphics[width=\linewidth]{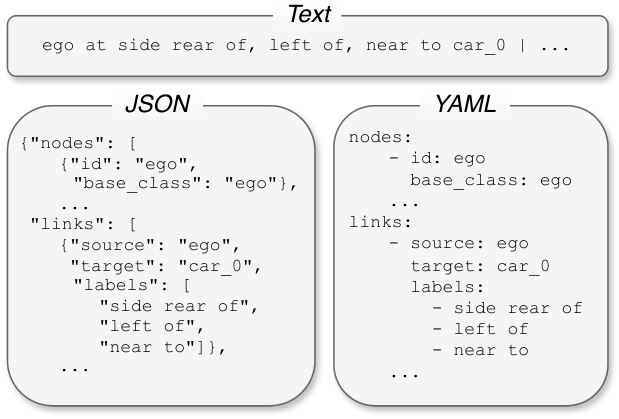}
    \caption{\textbf{Serialization formats.} We serialize scene graphs as Text, JSON, or YAML, each encoding subject-predicate-object triplets. 
    }
    \label{fig:scene_graph_serialization}
  \end{minipage}
  \hfill 
  \begin{minipage}[b]{0.61\linewidth}
    \centering
    \includegraphics[width=\linewidth]{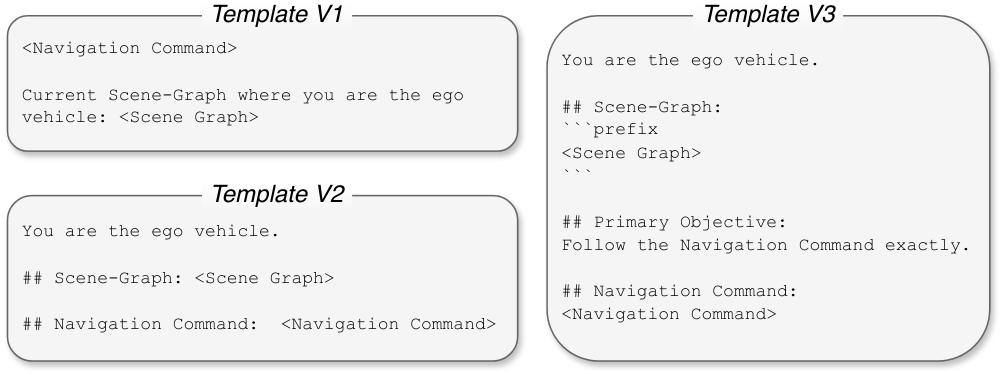}
    \caption{\textbf{Prompt templates.} Three prompt templates combine scene graphs with navigation commands: V1 uses direct concatenation, V2 adds ego-role framing and section headers, and V3 introduces a structured preamble with markdown-style fencing for consistent formatting.}
    \label{fig:scene_graph_injection}
  \end{minipage}
  \vspace{-0.5em}
\end{figure*}

\subsection{Scene Graph Serialization}
We serialize each graph as subject-predicate-object statements using a fixed, human-readable relation vocabulary. \cref{fig:scene_graph_serialization} illustrates the three formats we use, i.e., Text, JSON, and YAML, and a concrete example for each. To avoid redundancy, multiple predicates between the same ordered pair \((s,o)\) are emitted once as a multi-label statement rather than as duplicated links. When linearized, we list statements in hierarchical order with roads to junctions, lanes to roads, lane to lane connectivity, objects to lanes, and actors to lanes, and finally interactions between actors.

\noindent \textbf{Text.}
The Text form is a compact, linearized sequence of statements separated by a vertical bar (\texttt{|}). It applies two packing rules: \textit{Membership grouping} aggregates multiple sources that share the same hierarchical target, yielding “$s_1, s_2$ \textit{is in} $o$”. \textit{Predicate merging} collects all predicates that hold for a fixed ordered pair, yielding “$s$ $p_1, p_2, \ldots$ $o$”. Statements are emitted in a fixed order with hierarchical assignments first and actor-actor interactions following. 

\noindent \textbf{Structured Formats (JSON/YAML).} 
Both formats mirror the graph using two collections: \textit{nodes} (storing \textit{id} and \textit{base\_class}) and \textit{links}. To ensure a compact schema, each link represents a unique ordered pair \((s,o)\) and aggregates all applicable predicates into a single \textit{labels} list. While JSON provides a standard parser-friendly structure, YAML uses indentation-based nesting, which yields a more token-efficient representation by omitting braces and quotes.

\subsection{Scene Graph Injection}

We instantiate three prompt templates that differ in layout and instruction framing while keeping the underlying content identical. \cref{fig:scene_graph_injection} shows all three templates side by side for direct comparison. 
The first template (\textit{V1}) concatenates the navigation command and the scene graph in a single line without section headers. The command appears first, followed by a short cue and the serialized graph, minimizing prompt overhead and matching the instruction-plus-context compositions observed in prior work.
The second template (\textit{V2}) introduces explicit section headers for the scene graph and for the navigation command and begins with a reminder that the model acts as the ego vehicle. The graph is shown as a headed block and the command as a separate headed block, which makes both inputs distinct and separable when contexts become longer.
The third template (\textit{V3}) keeps the sectioning and wraps the scene graph in a fenced block with a format prefix such as \texttt{text}, \texttt{json}, or \texttt{yaml}. It also states a short \textit{Primary Objective} that emphasizes adherence to the external navigation command. The fence treats the graph as read-only and aims to stabilize tokenization for larger graphs or mixed formatting, and the format prefix encourages consistent parsing across representations.

\subsection{Scene Graph Conditioning}
We condition the model on structured relational information by including a serialized scene graph alongside the natural-language navigation command in the input prompt. In this work, we use the term \emph{conditioning} to refer specifically to prompt-level supervision: the model receives explicit relational context as part of its input, but its architecture, training objective, and loss functions remain unchanged. This prompt-based conditioning allows the model to ground decision-making and planning in spatial topology, regulatory context, and inter-actor interactions that would otherwise need to be inferred implicitly from visual features alone. Since scene graph-conditioning operates purely at the input level, it is model-agnostic and does not require modifying perception modules, introducing graph encoders, or defining auxiliary training losses.

\section{Experiments}
\label{sec:experiments}

\subsection{Benchmarks and Metrics}
We evaluate our approach using closed-loop simulation on two CARLA-based~\cite{dosovitskiy2017carla} benchmarks: LangAuto~\cite{shao2024lmdrive}, which replaces discrete global goals with natural-language navigation instructions across eight diverse towns, and Bench2Drive~\cite{jia2024bench2drive}, which features 44 interactive scenarios distributed across 15 towns and 220 short routes.
We report the official Driving Score (DS), defined as the product of Route Completion (RC) and Infraction Score (IS), alongside the Success Rate (SR) for Bench2Drive, with results averaged across three execution seeds.
For real-world open-loop evaluation, we additionally use nuScenes~\cite{caesar2020nuscenes}, reporting trajectory L2 error, collision rate, and intersection rate, measuring waypoint accuracy, collisions with dynamic agents, and incursions into non-drivable areas, respectively.

\subsection{Baselines and Data}
We evaluate our approach across four distinct vision-language baselines to demonstrate architectural generalizability: LMDrive~\cite{shao2024lmdrive} (LLaVA-v1.5~\cite{liu2024llava}) and BEVDriver~\cite{winter2025bevdriver} (LLaMA-7B~\cite{touvron2023llama}) on LangAuto, SimLingo~\cite{renz2025simlingo} (InternVL2-1B~\cite{gao2024mini}) on Bench2Drive, and SpaceDrive~\cite{li2026spacedrive} (Qwen2.5-VL-7B~\cite{bai2025qwen25vltechnicalreport}) on nuScenes.

\noindent \textbf{Dataset Collection.}
Since the original datasets lack relational annotations, we augment them with programmatically generated scene graphs.
For LMDrive and SimLingo, we extract graphs from privileged simulator states using 3D bounding boxes and HD map metadata. 
For nuScenes, we derive Actor-Only graphs directly from annotated 3D bounding-box poses.
We use the extended LMDrive data to train LMDrive and BEVDriver, the SimLingo data for SimLingo, and nuScenes for SpaceDrive.

\subsection{System Configurations}
For our approach, we define four scene graph usage settings, i.e., SG00, SG01, SG10, and SG11, to evaluate the impact of relational supervision on driving performance. 
SG00 uses no scene graphs during training or test-time and serves as our main baseline. We distinguish between two variants:
SG00-PT uses the original pretrained checkpoints provided by the model authors, while SG00-FT further fine-tunes these checkpoints on the corresponding training data.
This distinction ensures a fair comparison, as directly comparing our scene graph-conditioned models to SG00-PT would conflate the effects of additional training data with those of scene graph supervision.
SG01 includes scene graphs only at test-time. This setting probes whether pretrained models can benefit from relational structure without any prior exposure to scene graphs during training.
SG10 includes scene graphs during training but omits them at test-time. This allows to measure whether models can internalize relational priors and generalize without needing explicit scene graph input during inference.
SG11 includes scene graphs at both training and test-time and represents the fully supervised system configuration.

\subsection{Implementation and Training}
We initialize all models from officially released checkpoints and follow their respective optimization protocols.
The trainable components differ across architectures.
LMDrive freezes both the vision encoder and language model while optimizing the Q-Former~\cite{li2023blip}.
BEVDriver freezes the vision encoder but trains the language model via LoRA~\cite{hu2022lora} alongside the Q-Former.
SimLingo utilizes a fully trainable vision encoder paired with a LoRA-optimized language backbone.
SpaceDrive fine-tunes the core language model using LoRA while keeping the vision encoder and vision-language projector frozen.

For the CARLA-based experiments, LMDrive and BEVDriver are trained for five epochs, while SimLingo is fine-tuned for two epochs, using eight NVIDIA L40 GPUs.
For SpaceDrive, we fine-tune the official checkpoint for two additional epochs on nuScenes.
Scene graphs for the SG10 and SG11 settings are injected directly into the prompt without introducing any architectural modifications or auxiliary training losses.     
Training time generally increases with the token footprint of the chosen serialization format, making the compact Text representation more efficient than verbose JSON or YAML structures, while the choice of injection template has comparatively little impact on runtime (see Supplementary Material for detailed training duration).

\subsection{Results and Ablations}

\begin{table*}[!t]

  \caption{\textbf{Zero-shot test-time scene graph injection (SG01).} Pretrained LMDrive performance across abstraction levels, formats, and prompt templates, averaged over all LangAuto tracks. The rightmost column shows mean token counts.
  }
  \label{tab:prompt_config}
  \centering
  \footnotesize
  \begin{tabular}{c l *{4}{ccc} c}
    \toprule
    Serialization & Abstraction & 
      \multicolumn{3}{c}{V1} &
      \multicolumn{3}{c}{V2} &
      \multicolumn{3}{c}{V3} &
      \multicolumn{3}{c}{Mean} &
      Tokens \\
    \cmidrule(lr){3-5}\cmidrule(lr){6-8}\cmidrule(lr){9-11}\cmidrule(lr){12-14}
    & & DS $\uparrow$ & RC $\uparrow$ & IS $\uparrow$ & DS $\uparrow$ & RC $\uparrow$ & IS $\uparrow$ & DS $\uparrow$ & RC $\uparrow$ & IS $\uparrow$ & DS $\uparrow$ & RC $\uparrow$ & IS $\uparrow$ &  \\
    \midrule
    \multirow{3}{*}{Free-form}
      & Full        & 37.3 & 44.9 & 0.85 & 39.7 & 48.7 & 0.84 & 41.5 & 50.2 & 0.84 & 39.5 & 47.9 & 0.84 & 500 \\
      & Road-Level & 36.0 & 44.7 & 0.83 & 41.3 & 50.0 & 0.84 & 42.3 & 51.9 & 0.81 & 39.9 & 48.9 & 0.83 & 426 \\
      & Actor-Only & 41.0 & 48.8 & 0.85 & 41.7 & 50.4 & 0.82 & 41.0 & 49.7 & 0.83 & 41.2 & 49.6 & 0.83 & 106 \\
    \midrule
    \midrule
    \multirow{3}{*}{Text}
      & Full       & 38.9 & 46.2 & 0.84 & 41.3 & 49.7 & 0.83 & 41.8 & 49.4 & 0.83 & 40.7 & 48.4 & 0.83 & 437 \\
      & Road-Level & 40.6 & 47.0 & 0.86 & 40.5 & 50.1 & 0.81 & 41.5 & 49.4 & 0.83 & 40.9 & 48.8 & 0.83 & 370 \\
      \rowcolor{gray!15} %
      & Actor-Only & 40.0 & 48.7 & 0.85 & 43.1 & 51.2 & 0.84 & 44.7 & 53.0 & 0.84 & \textbf{42.6} & 51.0 & 0.84 & 69 \\
    \midrule
    \multirow{3}{*}{JSON}
      & Full       & 34.6 & 42.2 & 0.84 & 41.5 & 48.7 & 0.84 & 44.6 & 54.6 & 0.83 & 40.2 & 48.5 & 0.84 & 2905 \\
      & Road-Level & 34.4 & 43.4 & 0.83 & 41.8 & 51.2 & 0.81 & 46.6 & 56.3 & 0.82 & 41.0 & 50.3 & 0.82 & 2172 \\
      \rowcolor{gray!15} %
      & Actor-Only & 40.3 & 47.3 & 0.85 & 42.6 & 52.3 & 0.82 & 44.7 & 55.6 & 0.82 & \textbf{42.6} & 51.7 & 0.83 & 409 \\
    \midrule
    \multirow{3}{*}{YAML}
      & Full       & 39.4 & 45.5 & 0.85 & 41.8 & 50.4 & 0.82 & 41.8 & 51.9 & 0.80 & 41.0 & 49.3 & 0.82 & 1712 \\
      & Road-Level & 39.2 & 46.6 & 0.85 & 41.8 & 51.5 & 0.82 & 42.1 & 50.6 & 0.83 & 41.1 & 49.6 & 0.83 & 1246 \\
      \rowcolor{gray!15} %
      & Actor-Only & 42.8 & 52.0 & 0.84 & 43.6 & 53.2 & 0.82 & 44.7 & 52.7 & 0.84 & \textbf{43.7} & 52.6 & 0.83 & 242 \\
    \midrule
    \multirow{3}{*}{Mean}
      & Full
      & 37.6 & 44.6 & 0.84
      & 41.5 & 49.6 & 0.83
      & 42.7 & 52.0 & 0.82
      & 40.6 & 48.3 & 0.83 & 1685 \\
      & Road-Level
      & 38.1 & 45.7 & 0.84
      & 41.4 & 50.9 & 0.81
      & 43.4 & 52.1 & 0.83
      & 41.0 & 49.6 & 0.83 & 928 \\
      \rowcolor{gray!15} %
      & Actor-Only
      & \textbf{41.1} & 49.4 & 0.85
      & \textbf{43.1} & 52.2 & 0.82
      & \textbf{44.7} & 53.8 & 0.83
      & \textbf{43.0} & 51.8 & 0.84 & 240 \\
    \bottomrule
  \end{tabular}
\end{table*}

\begin{figure*}[!t]
    \centering
    \includegraphics[width=\textwidth]{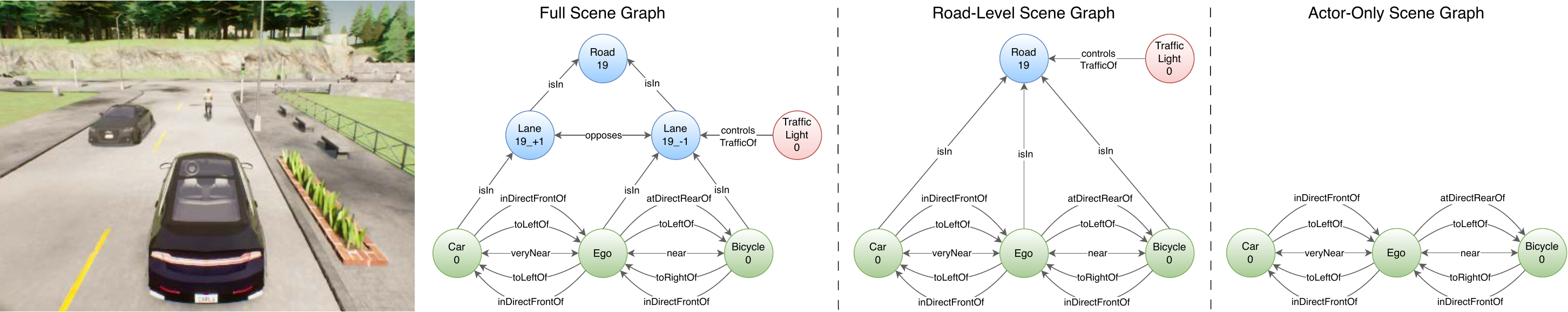}
    \caption{\textbf{Scene graph abstractions.} Comparison of Full, Road-Level, and Actor-Only abstractions, demonstrating how structural detail is progressively filtered.}
    \label{fig:scene_graph_example}

    \vspace{-0.5em}
\end{figure*}

\noindent \textbf{Zero-Shot Test-Time Scene Graph Injection (SG01).}
We begin by evaluating which scene graph integration strategies are most effective when applied at test-time only, without additional training, to assess whether a pretrained model can benefit from relational context without prior explicit conditioning.
This analysis characterizes the effects of abstraction level, serialization format, and injection template and motivates the configurations studied subsequently.
Specifically, we evaluate all combinations of abstraction level, serialization format, and injection template. \cref{tab:prompt_config} reports DS, RC, and IS for LMDrive averaged over the LangAuto \textit{Tiny}, \textit{Short}, and \textit{Long} tracks.
We provide mean rows and columns to facilitate comparisons across serialization formats and injection templates. 
Notably, multiple test-time configurations already surpass the performance of the original pretrained baselines, underscoring the inherent potential of relational grounding even without fine-tuning.
Three consistent trends emerge from this zero-shot evaluation.
First, structured formats (Text, JSON, YAML) consistently outperform unstructured \textit{Free-form} prose describing the same underlying relational elements as an image-caption-style narrative across all abstractions.
Second, injection template V3 generally performs best across configurations, while V1 tends to underperform the more structured V2 and V3 templates.
Third, Actor-Only abstractions yield the best results across all formats, indicating that minimal but structured relational information is most beneficial for pretrained models.
Thus, Actor-Only graphs optimize the trade-off between semantic relevance and token footprint, see \cref{fig:scene_graph_example}. 
These results motivate our focus on compact Actor-Only graphs and templates V2 and V3 in the subsequent SG10 experiments. We independently evaluate the abstraction choice under controlled SG10 training in \cref{tab:sg10_analysis}.

\noindent \textbf{Knowledge Internalization (SG10).}
\cref{tab:sg_augmented_ft} reports the driving performance for our SG10 configurations using the optimal Actor-Only abstraction and injection templates (V2, V3). To ensure fair, environment-matched comparisons, we reproduce all results on our local hardware using the original pretrained baselines (SG00-PT) and models fine-tuned without scene graphs (SG00-FT). This setup allows us to assess the extent to which structured relational context is internalized during training to benefit test-time, where graphs are absent.

\begin{table*}[!t]
  \caption{\textbf{Knowledge internalization (SG10).} Performance of our SG10 strategy across LMDrive, BEVDriver, and SimLingo using Actor-Only graphs compared to pretrained (SG00-PT) and fine-tuned (SG00-FT) baselines. DS is reported as mean $\pm$ standard deviation over three execution seeds.}
  \label{tab:sg_augmented_ft}
  \centering
  \fontsize{7pt}{7pt}\selectfont
  \setlength{\tabcolsep}{4.0pt}
  \begin{tabular}{c *{4}{ccc} cc cc}
    \toprule
    Config. &
      \multicolumn{6}{c}{LMDrive} &
      \multicolumn{6}{c}{BEVDriver} & \multicolumn{4}{c}{SimLingo} \\
    \cmidrule(lr){2-7}\cmidrule(l){8-13}\cmidrule(l){14-17}
    & \multicolumn{3}{c}{V2} & \multicolumn{3}{c}{V3} &
        \multicolumn{3}{c}{V2} & \multicolumn{3}{c}{V3} & \multicolumn{2}{c}{V2} & \multicolumn{2}{c}{V3} \\
    \cmidrule(l){2-4}\cmidrule(l){5-7}\cmidrule(l){8-10}\cmidrule(l){11-13}\cmidrule(l){14-15}\cmidrule(l){16-17}
    & DS $\uparrow$ & RC $\uparrow$ & IS $\uparrow$ & DS $\uparrow$ & RC $\uparrow$ & IS $\uparrow$ & DS $\uparrow$ & RC $\uparrow$ & IS $\uparrow$ & DS $\uparrow$ & RC $\uparrow$ & IS $\uparrow$ & DS $\uparrow$ & SR $\uparrow$ & DS $\uparrow$ & SR $\uparrow$ \\
    \midrule
    SG00-PT
      & 42.9 $\pm$ 2.3 & 54.8 & 0.80 & 42.9 $\pm$ 2.3 & 54.8 & 0.80
      & 44.7 $\pm$ 2.0 & 49.7 & 0.90 & 44.7 $\pm$ 2.0 & 49.7 & 0.90 
      & 87.4 $\pm$ 0.5 & 0.72 & 87.4 $\pm$ 0.5 & 0.72 \\
    SG00-FT
      & 41.5 $\pm$ 2.0 & 48.2 & 0.88 & 41.5 $\pm$ 2.0 & 48.2 & 0.88
      & 48.9 $\pm$ 2.3 & 56.4 & 0.86 & 48.9 $\pm$ 2.3 & 56.4 & 0.86 
      & 87.4 $\pm$ 0.5 & 0.71 & 87.4 $\pm$ 0.5 & 0.71 \\
    \midrule
    Text
      & 46.8 $\pm$ 1.7 & 56.2 & 0.84 & \textbf{51.8} $\pm$ 1.8 & 59.4 & 0.86
      & 51.7 $\pm$ 1.5 & 59.8 & 0.87 & 51.5 $\pm$ 1.8 & 60.0 & 0.86 
      & 88.4 $\pm$ 0.7 & 0.72 & 88.1 $\pm$ 0.7 & 0.71 \\
    JSON
      & 45.7 $\pm$ 2.1 & 53.5 & 0.85 & 45.5 $\pm$ 2.7 & 54.0 & 0.85
      & 47.3 $\pm$ 1.9 & 52.1 & 0.90 & 42.2 $\pm$ 2.9 & 50.7 & 0.85 
      & \textbf{88.9} $\pm$ 0.4 & 0.73 & 88.4 $\pm$ 0.2 & 0.72 \\
    YAML
      & 47.9 $\pm$ 1.7 & 57.4 & 0.83 & 46.8 $\pm$ 2.3 & 54.8 & 0.84
      & \textbf{56.1} $\pm$ 1.0 & 64.1 & 0.87 & 48.2 $\pm$ 1.4 & 55.7 & 0.85 
      & 88.2 $\pm$ 0.5 & 0.71 & 87.9 $\pm$ 0.2 & 0.70 \\
    \midrule
    \rowcolor{gray!15} %
    Mean
      & 46.8 & 55.7 & 0.84 & \textbf{48.0} & 56.1 & 0.85
      & \textbf{51.7} & 58.7 & 0.88 & 47.3 & 55.5 & 0.85 
      & \textbf{88.5} & 0.72 & 88.2 & 0.71 \\
    \bottomrule
  \end{tabular}
  
  \vspace{-0.5em}
  
\end{table*}

\begin{figure*}[t!]
    \centering
    \includegraphics[width=\textwidth]{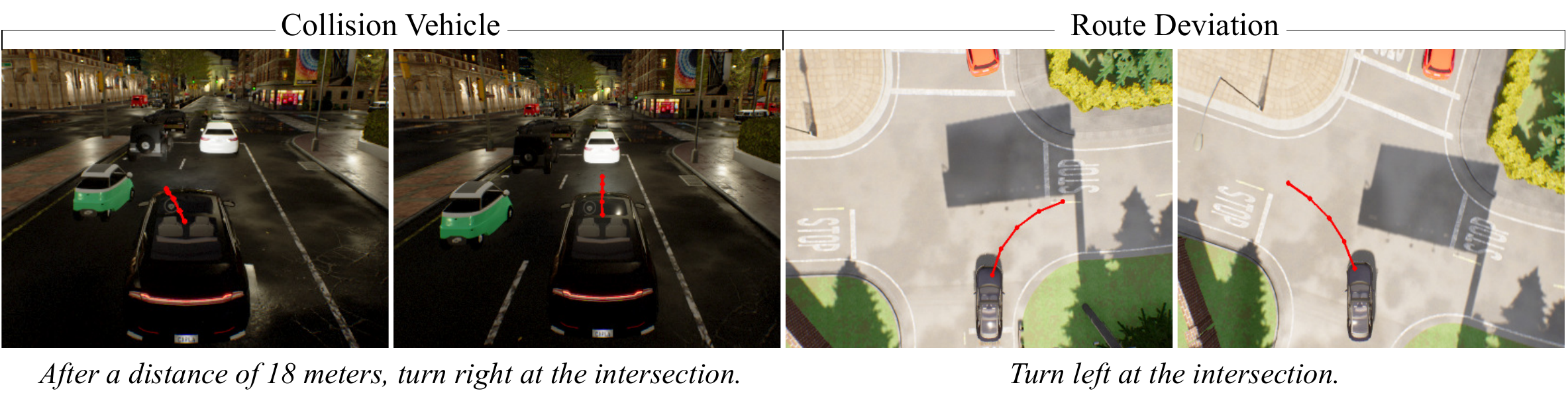}
    \caption{\textbf{Qualitative behavior comparison.} In each scenario, the LMDrive baseline (SG00-PT) on the left is compared against our SG10 on the right. The baseline predicts an unsafe lane change into traffic while failing to anticipate a right turn (Left Pair), and incorrectly turns right instead of left, whereas subsequently entering the opposite lane (Right Pair). In both scenarios, SG10 demonstrates superior command grounding and relational awareness.}
    \label{fig:qualitative_results}
    
    \vspace{-0.5em}
    
\end{figure*}

For LMDrive, our SG10 approach consistently outperforms both the pretrained baseline and the standard fine-tuning setup. 
The best configuration (Text, V3) achieves a DS of $51.8$, marking a substantial improvement over the baseline ($42.9$) and the SG00-FT variant ($41.5$). 
Notably, these results substantially surpass the best zero-shot SG01 performance (44.7) previously shown in \cref{tab:prompt_config}, demonstrating that explicit training with relational supervision is more effective than simple test-time injection.
BEVDriver shows a similar trend of substantial improvement. 
The top configuration (YAML, V2) reaches a DS of $56.1$, outperforming its respective SG00-PT ($44.7$) and SG00-FT ($48.9$) benchmarks. 
While BEVDriver shows a preference for YAML serialization and the V2 template, the underlying pattern remains consistent with LMDrive: training-time scene graph exposure enhances internal planning behavior even when that structure is removed during deployment.
To demonstrate that these benefits are not specific to the LangAuto benchmark, we further evaluate our strategy using SimLingo on the Bench2Drive dataset. 
Despite a high baseline performance of $87.4$, SimLingo achieves consistent improvements across all formats, with the best configuration (JSON, V2) reaching a DS of $88.9$. 
Across architectures, these persistent gains suggest that training-time scene graph supervision shapes internal representations toward relational structure, an interpretation we examine directly through controlled ablations and hidden-state probing below.
This SG10 strategy offers major practical advantages: it eliminates complex, potentially error-prone test-time scene graph generation and minimizes prompt overhead. 
By dropping the explicit scene graph during inference, the model completely avoids test-time token costs, matching the baseline's exact inference latency and low memory footprint.

\begin{table*}[t]
\centering
\caption{\textbf{Controlled analysis of SG10 and relational internalization.}
(a) Prompt-content ablation isolates the effects of prompt framing, entity information, and relational supervision.
(b) Abstraction-level comparison evaluates Full, Road-Level, and Actor-Only graphs under the same LMDrive Text/V3 SG10 configuration.
(c) Balanced accuracy (\%) of probes recovering 11 scene relations from graph-free LMDrive hidden states after SG10 training with different serialization formats.
DS is reported as mean $\pm$ standard deviation over three execution seeds.}
\label{tab:sg10_analysis}

\footnotesize
\setlength{\tabcolsep}{3pt}

\begin{minipage}[t]{0.31\textwidth}
\centering
(a) Prompt-content ablation

\vspace{0.3em}

\begin{tabular}{lccc}
\toprule
Graph Content & DS $\uparrow$ & RC $\uparrow$ & IS $\uparrow$ \\
\midrule
Empty graph & 44.1 $\pm$ 2.0 & 51.7 & \textbf{0.86} \\
No relations & 46.0 $\pm$ 4.3 & 52.5 & \textbf{0.86} \\
\rowcolor{gray!15}
Actor-Only & \textbf{51.8} $\pm$ 1.8 & \textbf{59.4} & \textbf{0.86} \\
\bottomrule
\end{tabular}
\end{minipage}
\hfill
%
\begin{minipage}[t]{0.29\textwidth}
\centering
(b) Scene-graph abstraction

\vspace{0.3em}

\begin{tabular}{lccc}
\toprule
Abstraction & DS $\uparrow$ & RC $\uparrow$ & IS $\uparrow$ \\
\midrule
Full & 43.6 $\pm$ 2.8 & 50.8 & 0.85 \\
Road-Level & 46.2 $\pm$ 3.4 & 52.9 & \textbf{0.87} \\
\rowcolor{gray!15}
Actor-Only & \textbf{51.8} $\pm$ 1.8 & \textbf{59.4} & 0.86 \\
\bottomrule
\end{tabular}
\end{minipage}
\hfill
%
\begin{minipage}[t]{0.36\textwidth}
\centering
(c) Relational information probing

\vspace{0.3em}

\begin{tabular}{lcccc}
\toprule
Probe & SG00-FT & JSON & Text & YAML \\
\midrule
Linear & 77.2 & 77.8 & 77.9 & \textbf{78.2} \\
MLP-128 & 77.7 & 80.2 & 78.8 & \textbf{80.3} \\
MLP-256 & 77.1 & 79.3 & 79.3 & \textbf{79.8} \\
\bottomrule
\end{tabular}
\end{minipage}

\vspace{-0.25em}
\end{table*}

\begin{table}[t]
\centering
\caption{\textbf{Transfer to real-world nuScenes data.}
Open-loop SpaceDrive planning performance after two additional fine-tuning epochs from the official checkpoint, without (SG00-FT) and with (SG10) scene-graph supervision using text serialization.
Overall denotes the mean across metrics.}
\label{tab:spacedrive_nuscenes}
\footnotesize
\begin{tabular}{lcccc}
\toprule
Method & L2 $\downarrow$ & Collision $\downarrow$
& Intersection $\downarrow$ & Overall $\downarrow$ \\
\midrule
SG00-PT & 0.321 & 0.228 & 1.315 & 0.621 \\
SG00-FT & 0.332 & \textbf{0.195} & 1.446 & 0.658 \\
\rowcolor{gray!15}
SG10, Text & \textbf{0.325} & 0.227 & \textbf{1.192} & \textbf{0.581} \\
\bottomrule
\end{tabular}


\end{table}

\noindent \textbf{Disentangling SG10 Gains.}
While the preceding results demonstrate substantial improvements from SG10, they do not isolate whether these gains originate from relational supervision itself or from auxiliary effects such as the modified prompt structure and additional semantic content.
We therefore conduct controlled experiments with LMDrive while fixing the Text serialization and V3 prompt template (\cref{tab:sg10_analysis}a).
Providing the V3 prompt with an empty scene graph already improves DS from $41.5$ for SG00-FT to $44.1$, indicating a modest benefit from the prompt structure alone.
Retaining the graph entities while removing all relations further increases DS to $46.0$.
Importantly, restoring the complete relational content raises DS substantially to $51.8$, corresponding to a further gain of $+5.8$ DS over the relation-free variant.
This improvement is primarily reflected in higher route completion ($52.5 \rightarrow 59.4$), while the infraction score remains unchanged ($0.86$), indicating that explicit relations particularly benefit task progress and navigation rather than merely altering the safety penalty.

We additionally evaluate whether the preference for Actor-Only graphs observed in the zero-shot SG01 study persists under controlled SG10 training (\cref{tab:sg10_analysis}b).
Keeping the model, serialization, and prompt template fixed, performance increases from $43.6$ DS with Full graphs to $46.2$ with Road-Level graphs and $51.8$ with Actor-Only graphs.
Thus, the advantage of Actor-Only representations is retained independently under SG10 training and is not restricted to the configuration selection observed in SG01.
Together, these experiments show that prompt framing and entity information explain part of the improvement, while relational supervision provides a substantial additional contribution to the SG10 gains.

\noindent \textbf{Probing Relational Internalization.}
The preceding ablations establish that relational content contributes substantially to the behavioral gains of SG10. We next investigate whether this improvement is accompanied by a stronger representation of scene relations within the model when scene graphs are absent at inference.
To this end, we extract the LLM hidden-state representations of LMDrive under graph-free inference and train lightweight probes to predict the presence of spatial and interaction relations defined in \cref{fig:scene_graph_construction}.
We evaluate both a linear probe and nonlinear MLP probes with hidden dimensions of 128 and 256, reporting macro-averaged balanced accuracy across all relations in \cref{tab:sg10_analysis}c.
Across all probe architectures and scene-graph serialization formats used during SG10 training, relational information is more decodable than for SG00-FT.
For the linear probe, balanced accuracy increases from $77.2\%$ for SG00-FT to $77.8$--$78.2\%$ after SG10 training, showing that the effect is already present under a low-capacity readout.
The improvement becomes more pronounced with nonlinear probes: MLP-128 reaches up to $80.3\%$ compared with $77.7\%$ for SG00-FT, while MLP-256 improves from $77.1\%$ to as much as $79.8\%$.
These results provide representation-level evidence that scene-graph-conditioned training increases the accessibility of relational information in the graph-free hidden states, complementing the behavioral improvements observed above.

\noindent \textbf{Real-World Transfer.}
Finally, we investigate whether the benefits of scene graph supervision transfer beyond CARLA-based closed-loop simulation to real-world driving data.
We evaluate the SG10 paradigm with SpaceDrive on nuScenes (\cref{tab:spacedrive_nuscenes}).
Compared with the corresponding SG00-FT model, SG10 improves the average L2 error from $0.332$ to $0.325$ and substantially reduces the intersection metric from $1.446$ to $1.192$, resulting in an overall improvement from $0.658$ to $0.581$.
While the collision metric increases from $0.195$ to $0.227$, it remains comparable to the released checkpoint ($0.228$).
Notably, SG10 also improves the overall metric over the original checkpoint ($0.621 \rightarrow 0.581$), despite a marginally higher L2 error ($0.321 \rightarrow 0.325$).
These results demonstrate that the benefits of training-time relational supervision transfer to a different language-based planning architecture and real-world nuScenes data.

\noindent \textbf{Additional Analysis.}
Beyond our primary SG10 analysis, we provide several complementary experiments in the Supplementary Material, including an extended breakdown of driving infractions, explicit relational conditioning with scene graphs retained at inference (SG11), and robustness to increasing levels of scene-graph corruption.
We additionally report computational overhead, including graph processing, inference latency, and memory usage.

\section{Conclusion}
\label{sec:conclusion}

We presented a model-agnostic training paradigm that grounds language-based autonomous driving models in explicit relational structure through serialized traffic scene graphs. 
By systematically evaluating different abstraction levels, serialization formats, and injection templates, we identified lightweight and effective strategies that improve driving performance without requiring architectural changes or additional training objectives. 
Among all configurations, Actor-Only abstractions provide a strong trade-off between semantic relevance and token footprint by capturing essential dynamic interactions with minimal overhead.
While test-time prompts (SG01) enhance pretrained models, our primary internalization paradigm (SG10) utilizes scene graphs strictly during training to yield substantial and consistent performance gains across diverse baseline architectures. 
Controlled ablations and representation probing further show that relational information contributes substantially to these gains and becomes more decodable under graph-free inference.
Moreover, evaluation with SpaceDrive on nuScenes demonstrates that these benefits transfer beyond CARLA simulation to real-world driving data.
By omitting scene graphs during deployment, SG10 avoids test-time graph-generation errors and additional token overhead while retaining the benefits of relational supervision.

\noindent \textbf{Limitations.}
GraphPilot currently represents scene graphs through textual serialization, which is model-agnostic but token-inefficient and sensitive to serialization choices such as node, edge, and predicate ordering. Although we use a deterministic ordering, semantically equivalent graphs may still induce different language representations. Future work could employ graph-native encoders, such as GNNs, to better exploit graph topology, learn task-relevant entity and relation weighting, and project a compact graph representation directly into the language embedding space.

\noindent \textbf{Acknowledgments.} This work has been funded by the BMFTR under support code 13FH544KB2 (ADRIVE-GPT). The authors alone are responsible for the content of the paper. The authors thank the DACHS data analysis cluster, hosted at Esslingen University and co-funded by the MWK within the DFG's "Großgeräte der Länder" program, for providing the computational resources necessary for this research.

{
    \small
    \bibliographystyle{ieeenat_fullname}
    \bibliography{main}
}

\clearpage
\maketitlesupplementary

This supplementary material provides additional information to support the results presented in the main paper. 
We include detailed implementation settings, full training-time measurements, extended quantitative results, and additional qualitative analyses. 
All supplementary content is intended to enhance reproducibility and provide full transparency into our experimental pipeline.

\section{Implementation and Reproducibility}
\label{sec:implementation_details}

This section provides additional implementation details for all evaluated models, including optimization settings, dataset statistics, scene graph generation, and exemplary serialized inputs.
Unless stated otherwise, we follow the official training and optimization protocols of the respective baselines and initialize all models from their publicly released checkpoints.

\subsection{Model and Training Details}

\begin{table}[!b]
\caption{\textbf{Implementation Details for LMDrive and BEVDriver.}
Both methods use their best-performing language backbones with official pretrained checkpoints (LMDrive: LLaVA-v1.5, BEVDriver: LLaMA-7B). LoRA parameters apply only to the trainable language backbone of BEVDriver.}

\label{tab:impl_details_lmdrive_bevdriver}
\centering
\footnotesize
\begin{tabular}{l l}
\toprule
Category & Hyperparameter \\
\midrule

\multirow{5}{*}{Model} 
    & freeze\_vit: True \\
    & use\_notice\_prompt: False \\
    & LoRA rank: 16 \\
    & LoRA alpha: 32 \\
    & LoRA dropout: 0.05 \\

\midrule

\multirow{3}{*}{Dataset}
    & enable\_start\_frame\_augment: True \\
    & token\_max\_length: 40 \\
    & enable\_notice: False \\

\midrule

\multirow{10}{*}{Training}
    & Optimizer: AdamW \\
    & lr\_sched: linear\_warmup\_cosine\_lr \\
    & init\_lr: 1e-4 \\
    & min\_lr: 1e-5 \\
    & warmup\_lr: 1e-6 \\
    & warmup\_steps: 2000 \\
    & weight\_decay: 0.06 \\
    & max\_epoch: 5 \\
    & batch\_size\_train: 1 \\
    & world\_size: 8 \\

\bottomrule
\end{tabular}
\end{table}

\noindent\textbf{LMDrive and BEVDriver.}
LMDrive and BEVDriver share the same underlying training infrastructure and are fine-tuned on the same extended LMDrive training split.
Their optimization settings are summarized in \cref{tab:impl_details_lmdrive_bevdriver}.
Both models freeze the visual backbone, but differ in the trainable language components: LMDrive keeps the language model frozen and optimizes the Q-Former, whereas BEVDriver additionally adapts the language model using LoRA.
For all SG10 and SG11 experiments, the serialized scene graph is appended to the corresponding language prompt without modifying the model architecture, prediction heads, or training objectives.

\begin{table}[!t]
\caption{\textbf{Implementation Details for SimLingo.} Optimization strategy for fine-tuning the official pretrained checkpoint using InternVL2-1B as language backbone.}
\label{tab:impl_details_simlingo}
\centering
\footnotesize
\begin{tabular}{l l}
\toprule
Category & Hyperparameter \\
\midrule

\multirow{3}{*}{Model} 
    & LoRA rank: 16 \\
    & LoRA alpha: 16 \\
    & LoRA dropout: 0.05 \\

\midrule

\multirow{12}{*}{Dataset}
    & data\_module: carla\_bucket\_v12\_dreamer \\
    & pred\_len: 11 \\
    & use\_commentary: True \\
    & use\_qa: True \\
    & qa\_augmentation: True \\
    & img\_shift\_augmentation: False \\
    & hist\_len: 1 \\
    & route\_as: target\_point\_command \\
    & use\_lmdrive\_commands: True \\

\midrule

\multirow{9}{*}{Training}
    & Optimizer: AdamW \\
    & seed: 888 \\
    & lr\_sched: one\_cycle\_cosine \\
    & lr: 1e-6 \\
    & warmup\_steps: 5\% of total steps \\
    & weight\_decay: 0.2 \\
    & betas: (0.9, 0.999) \\
    & max\_epochs: 2 \\
    & batch\_size: 8 \\
    & gpus: 8 \\

\bottomrule
\end{tabular}
\end{table}

\noindent\textbf{SimLingo.}
SimLingo follows its original training pipeline and is initialized from the officially released InternVL2-1B checkpoint.
In contrast to LMDrive and BEVDriver, its vision encoder remains trainable while the language backbone is adapted using LoRA.
We retain the original input construction, route-command representation, commentary and question-answer supervision, and planning horizon.
The principal optimization settings are reported in \cref{tab:impl_details_simlingo}.
As for the LangAuto models, scene graph conditioning modifies only the textual input and introduces no additional loss functions or architectural components.

\noindent\textbf{SpaceDrive.}
For the nuScenes experiments, we initialize SpaceDrive from its officially released checkpoint, which uses Qwen2.5-VL-7B as its vision-language backbone and retain its original spatial encoding and planning configuration.
The vision encoder and vision-language projector remain frozen, while the core language model is adapted using LoRA with rank 16.
We use the camera-only planning configuration with $640\times640$ image resolution and retain both the visual-token and input/output 3D positional encodings.
Starting from the released checkpoint, SG00-FT and SG10 are fine-tuned for two additional epochs under identical optimization settings, differing only in the presence of scene graph supervision during training.
The corresponding configuration is summarized in \cref{tab:impl_details_spacedrive}.

\begin{table}[!t]
\caption{\textbf{Implementation Details for SpaceDrive.}
Optimization setup for fine-tuning the official SpaceDrive checkpoint with Qwen2.5-VL-7B.}
\label{tab:impl_details_spacedrive}
\centering
\footnotesize
\begin{tabular}{l l}
\toprule
Category & Hyperparameter \\
\midrule

\multirow{4}{*}{Model}
    & Backbone: Qwen2.5-VL-7B \\
    & LoRA rank: 16 \\
    & LoRA alpha: 64 \\
    & LoRA dropout: 0.3 \\
    & Vision 3D positional encoding: True \\
    & Input/output 3D positional encoding: True \\

\midrule

\multirow{5}{*}{Dataset}
    & Dataset: nuScenes \\
    & Input modality: Camera \\
    & Image resolution: $640\times640$ \\
    & Training task: Planning only \\
    & Sequential training: True \\

\midrule

\multirow{8}{*}{Training}
    & Optimizer: AdamW \\
    & lr\_sched: CosineAnnealing \\
    & lr: $5\times10^{-5}$ \\
    & warmup\_iters: 500 \\
    & weight\_decay: $10^{-4}$ \\
    & betas: $(0.9, 0.999)$ \\
    & additional\_epochs: 2 \\
    & batch\_size: $1\times8$ GPUs \\

\bottomrule
\end{tabular}
\end{table}

For the CARLA-based experiments, training is distributed across eight NVIDIA L40 GPUs.
Our compute nodes are equipped with AMD EPYC 9254 CPUs, 384~GB system memory, and 48~GB VRAM per GPU.
SpaceDrive is likewise trained in a distributed eight-GPU configuration with a per-GPU batch size of one.
Detailed training-time measurements for the different scene graph representations are provided in \cref{sec:training_time}.

\subsection{Dataset Statistics and Scene Graph Annotation}

\cref{tab:dataset_statistics} summarizes the amount of training data used for each baseline and the source from which scene graph annotations are derived.
For LMDrive and BEVDriver, we use 1,435,626 frames from 2,327 routes of the collected LMDrive training corpus, which contains 5,689,768 frames across 9,200 routes in total.
For SimLingo, we use all 1,565,612 frames from 5,192 training routes.
For SpaceDrive, we use all 34,149 annotated training keyframes from the 700 nuScenes training scenes.

For the CARLA-based datasets, scene graphs are generated programmatically from privileged simulator state using actor 3D bounding boxes and HD-map information.
These annotations provide the actor identities, roadway structure, and spatial relations required to construct the Full, Road-Level, and Actor-Only representations described in the main paper.
For nuScenes, we construct Actor-Only graphs directly from the annotated 3D bounding-box poses of traffic participants.
A corresponding scene graph is generated for every frame used during scene graph-conditioned training.
Thus, SG10 requires structured annotations only during training; no scene graph construction or privileged scene information is required at inference.

\begin{table}[!t]
\caption{\textbf{Dataset and scene graph statistics.}
Number of training and validation frames and routes/scenes used in our experiments, together with the source of scene graph annotations.}
\label{tab:dataset_statistics}
\centering
\footnotesize
\begin{tabular}{lccc}
\toprule
Dataset & Frames & Scenes & Scene Graph Source \\
\midrule
LMDrive
& 1,435,626 
& 2,327 
& CARLA state + HD map \\
SimLingo
& 1,565,612
& 5,192
& CARLA state + HD map \\
nuScenes
& 34,149
& 700
& Annotated 3D bounding boxes \\
\bottomrule
\end{tabular}
\end{table}

\subsection{Scene Graph and Prompt Examples}

\begin{figure*}[t!]
    \centering
    \includegraphics[width=\textwidth]{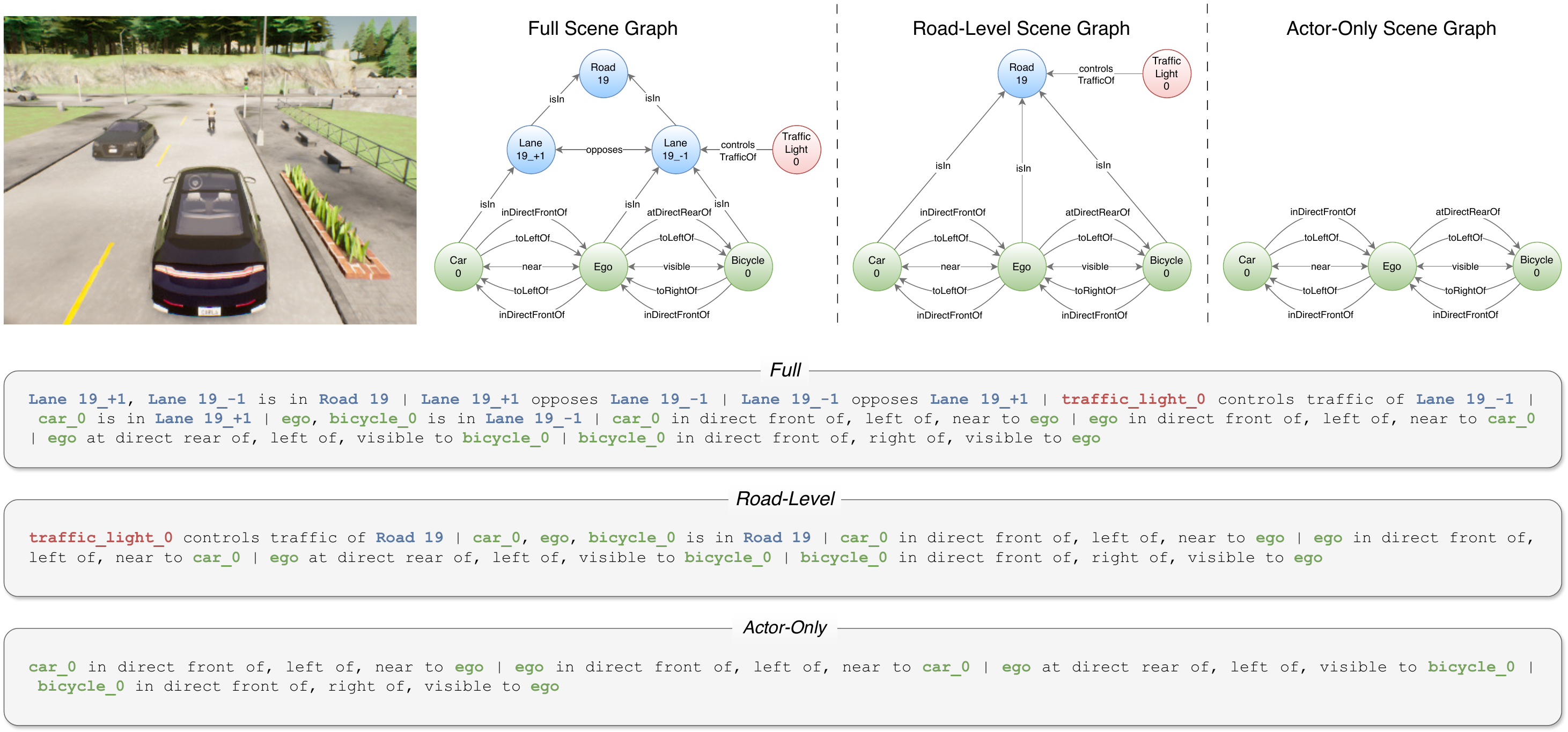}
    \caption{\textbf{Scene graph abstraction and serialization example.}
    A representative traffic scene is shown together with its Full, Road-Level, and Actor-Only graph representations and their corresponding Text serializations.
    Road-Level collapses lane-level structure into roads, while Actor-Only retains only traffic actors and their pairwise spatial relations.
    All three representations are derived from the same underlying scene.}
    \label{fig:sg_example_long}
\end{figure*}

To illustrate the exact relational information provided to the models, \cref{fig:sg_example_long} shows a representative CARLA scene together with its Full, Road-Level, and Actor-Only scene graphs and the corresponding Text serializations.
The Full representation retains roadway structure, traffic objects, actor memberships, and inter-actor relations.
Road-Level removes lane-level structural detail while preserving road-level memberships and regulatory context, whereas Actor-Only removes structural and traffic-object nodes entirely and retains only pairwise relations between traffic participants.
Importantly, these abstractions are derived from the same underlying scene and therefore differ only in the amount of structural information exposed to the language model.

The serialized strings shown in \cref{fig:sg_example_long} correspond to the actual Text representation used as model input.
For the JSON and YAML variants, the same nodes and relations are represented using the structured schemas introduced in the main paper.
Scene graphs are inserted into one of the three prompt templates together with the navigation command; SG10 uses this augmented prompt only during training and reverts to the original graph-free input format at inference.

\section{Computational Analysis}
\label{sec:computational_analysis}

\subsection{Training Time}
\label{sec:training_time}

\begin{table}[t!]
\caption{\textbf{Training time for LMDrive and BEVDriver.}
Elapsed training time in hours for SG10/SG11 fine-tuning across scene graph serialization formats, abstraction levels, and prompt templates.
Graph-free fine-tuning (SG00-FT) requires approximately $45$\,h for LMDrive and $20$\,h for BEVDriver.}
\label{tab:combined_training_time}
\centering
\setlength{\tabcolsep}{4pt}
\footnotesize
\begin{tabular}{@{}c c c ccc@{}}
\toprule
Model & Serialization & Template & Full & Road-Level & Actor-Only \\
\midrule
    \multirow{7}{*}{LMDrive} 
        & \multirow{2}{*}{Text} & V2 & 77\si{\hour} & 74\si{\hour} & 61\si{\hour} \\
        &  & V3 & 77\si{\hour} & 76\si{\hour} & 61\si{\hour} \\ \cmidrule(r){2-6}
    & \multirow{2}{*}{JSON} 
        & V2 & 107\si{\hour} & 105\si{\hour} & 77\si{\hour} \\
        &  & V3 & 107\si{\hour} & 106\si{\hour} & 78\si{\hour} \\ \cmidrule(r){2-6}
    & \multirow{2}{*}{YAML} 
        & V2 & 107\si{\hour} & 105\si{\hour} & 68\si{\hour} \\
    &  & V3 & 106\si{\hour} & 104\si{\hour} & 69\si{\hour} \\
\midrule
\multirow{7}{*}{BEVDriver} 
    & \multirow{2}{*}{Text} 
        & V2 & 51\si{\hour} & 48\si{\hour} & 34\si{\hour} \\
        &  & V3 & 60\si{\hour} & 58\si{\hour} & 35\si{\hour} \\ \cmidrule(r){2-6}
    & \multirow{2}{*}{JSON} 
        & V2 & 78\si{\hour} & 76\si{\hour} & 52\si{\hour} \\
        &  & V3 & 78\si{\hour} & 76\si{\hour} & 53\si{\hour} \\ \cmidrule(r){2-6}
    & \multirow{2}{*}{YAML} 
        & V2 & 77\si{\hour} & 76\si{\hour} & 43\si{\hour} \\
        &  & V3 & 77\si{\hour} & 76\si{\hour} & 44\si{\hour} \\
\bottomrule
\end{tabular}
\end{table}

\begin{table}[t!]
\caption{\textbf{Training time for SimLingo.}
Elapsed training time in hours across serialization formats and prompt templates using Actor-Only scene graphs.
Graph-free fine-tuning (SG00-FT) requires approximately $19$\,h.}
\label{tab:simlingo_training_time}
\centering
\footnotesize
\begin{tabular}{c c c}
\toprule
Serialization & Template 
  & Actor-Only \\
\midrule

\multirow{2}{*}{Text}
  & V2 & 20\si{\hour} \\
  & V3 & 21\si{\hour} \\
\midrule
\multirow{2}{*}{JSON}
  & V2 & 29\si{\hour} \\
  & V3 & 29\si{\hour} \\
\midrule
\multirow{2}{*}{YAML}
  & V2 & 21\si{\hour} \\
  & V3 & 21\si{\hour} \\

\bottomrule
\end{tabular}
\end{table}

We analyze the computational cost of scene graph-conditioned training across serialization formats, abstraction levels, and prompt templates.
For LMDrive and BEVDriver, we evaluate all combinations of Full, Road-Level, and Actor-Only graphs with Text, JSON, and YAML serialization using templates V2 and V3.
For SimLingo, we evaluate the same serialization and template variants using the Actor-Only abstraction employed in our main experiments.
We omit V1 from fine-tuning because the zero-shot analysis in the main paper consistently identifies V2 and V3 as the stronger prompt variants.

\cref{tab:combined_training_time,tab:simlingo_training_time} report the resulting training times.
For reference, graph-free fine-tuning (SG00-FT) requires approximately $45$\,h for LMDrive, $20$\,h for BEVDriver, and $19$\,h for SimLingo.
Scene graph conditioning increases training time primarily through the additional language-token sequence length.
Accordingly, Actor-Only representations are consistently more efficient than Full and Road-Level graphs, while the compact Text serialization generally incurs lower training cost than the more verbose JSON and YAML representations.
Differences between V2 and V3 are comparatively small, indicating that the scene graph content and serialization format dominate the additional computational cost rather than the prompt framing itself.

\begin{table*}[!ht]
\caption{\textbf{Test-time computational overhead.}
Latency (ms) and GPU memory usage (GB) for scene graph generation, abstraction, serialization, and LMDrive inference under SG11 using template V3.
Results are reported across abstraction levels and serialization formats as the 1st percentile (P01), median (Med), and 99th percentile (P99).
SG10 is shown as the graph-free inference reference and therefore incurs no scene graph processing or additional token overhead.}
\label{tab:sg_runtime}
  \centering
  \footnotesize
  \begin{tabular}{@{}cc ccc ccc ccc ccc ccc@{}}
    \toprule
    Serialization & Abstraction &
    \multicolumn{3}{c}{Generation [ms]} &
    \multicolumn{3}{c}{Abstraction [ms]} &
    \multicolumn{3}{c}{Serialization [ms]} &
    \multicolumn{3}{c}{Latency [ms]} &
    \multicolumn{3}{c}{GPU [GB]} \\
    \cmidrule(l){3-5}\cmidrule(l){6-8}\cmidrule(l){9-11}\cmidrule(l){12-14}\cmidrule(l){15-17} &
    & 
    P01 & Med & P99 &
    P01 & Med & P99 &
    P01 & Med & P99 &
    P01 & Med & P99 &
    P01 & Med & P99 \\
    \midrule
    \multirow{3}{*}{JSON}
      & Full        & 25.0 & 40.7 & 53.6 & 0.0 & 0.0 & 0.0 & 0.3 & 1.2 & 2.3 & 158.9 & 1259.2 & 2057.6 & 25.5 & 39.6 & 41.7 \\
      & Road-Level  & 25.8 & 38.6 & 52.8 & 0.1 & 0.2 & 1.1 & 0.4 & 0.9 & 1.9 & 203.9 & 1065.8 & 1967.4 & 23.7 & 36.0 & 41.9 \\
      & Actor-Only  & 25.6 & 39.2 & 54.6 & 0.1 & 0.2 & 1.0 & 0.8 & 1.3 & 2.6 & 288.3 & 351.1 & 516.4 & 23.1 & 26.2 & 29.3 \\
    \midrule
    \multirow{3}{*}{YAML}
      & Full        & 25.2 & 40.9 & 54.3 & 0.0 & 0.0 & 0.0 & 1.0 & 5.7 & 7.7 & 225.1 & 1056.6 & 1566.7 & 26.5 & 32.6 & 40.1 \\
      & Road-Level  & 25.7 & 41.4 & 54.7 & 0.1 & 0.2 & 0.9 & 1.1 & 4.8 & 7.5 & 287.4 & 924.5 & 1598.3 & 27.8 & 29.5 & 38.1 \\
      & Actor-Only  & 25.2 & 40.1 & 54.6 & 0.1 & 0.1 & 1.0 & 0.2 & 0.5 & 0.8 & 185.1 & 253.9 & 601.3 & 25.0 & 25.1 & 25.2 \\
    \midrule
    \multirow{3}{*}{Text}
      & Full        & 26.0 & 40.6 & 53.4 & 0.0 & 0.0 & 0.0 & 0.4 & 0.9 & 2.0 & 79.5 & 107.4 & 136.9 & 20.8 & 20.9 & 21.0 \\
      & Road-Level  & 25.3 & 41.4 & 54.0 & 0.1 & 0.2 & 0.4 & 0.2 & 0.8 & 1.6 & 58.1 & 93.3 & 132.3 & 19.6 & 19.6 & 19.8 \\
      & Actor-Only  & 25.4 & 38.7 & 53.6 & 0.1 & 0.1 & 0.3 & 0.2 & 0.5 & 0.7 & 46.4 & 85.4 & 118.9 & 18.2 & 18.7 & 18.9 \\
    \midrule
    \multicolumn{2}{c}{SG10} & 0.0 & 0.0 & 0.0 & 0.0 & 0.0 & 0.0 & 0.0 & 0.0 & 0.0 & 51.6 & 56.9 & 65.5 & 18.2 & 18.3 & 18.4 \\
    \bottomrule
  \end{tabular}
\end{table*}

Across LMDrive, BEVDriver, and SimLingo, we conducted 42 scene graph-conditioned fine-tuning runs: 18 each for LMDrive and BEVDriver and six for SimLingo.
The accumulated training time across these runs is 2,797\,h, comprising 1,568\,h for LMDrive, 1,088\,h for BEVDriver, and 141\,h for SimLingo.

\subsection{Test-Time Computational Overhead}
\label{sec:test_time_overhead}

We additionally quantify the computational overhead associated with scene graph processing and test-time conditioning.
All measurements are performed with LMDrive on a single NVIDIA L40S GPU.
For configurations that use scene graphs at inference, we separately measure graph generation, abstraction, serialization, model inference latency, and GPU memory consumption.
We evaluate all abstraction levels and serialization formats using template V3 and report the 1st percentile (P01), median, and 99th percentile (P99) in \cref{tab:sg_runtime}.

Scene graph generation itself is relatively stable across configurations, with a median latency of approximately $39$--$41$\,ms.
Abstraction and serialization contribute comparatively little additional processing time.
The dominant computational overhead instead arises during model inference, where the longer prompts associated with Full and Road-Level graphs, particularly in JSON and YAML, substantially increase both latency and GPU memory consumption.
Actor-Only representations reduce this overhead considerably, with Text providing the most computationally efficient serialization.

Importantly, our primary SG10 configuration does not require scene graph construction or additional prompt tokens at inference.
Its median inference latency is $56.9$\,ms with $18.3$\,GB GPU memory usage, closely matching the graph-free baseline behavior.
In contrast, retaining explicit scene graphs at inference can substantially increase runtime; for example, Full JSON reaches a median inference latency of $1259.2$\,ms and $39.6$\,GB memory usage.
These measurements highlight an additional practical advantage of SG10: relational supervision is introduced during training without incurring the test-time computational cost associated with explicit scene graph conditioning.

\section{Additional Experimental Analysis}
\label{sec:additional_analysis}

\subsection{Extended Infraction Analysis}

To better understand how scene graph-conditioned training affects closed-loop behavior, we report a detailed infraction breakdown for LMDrive in \cref{tab:is_analysis}.
Compared with SG00-PT, SG10 reduces vehicle collisions (CV) from $2.83$ to $1.17$ and layout collisions (CL) from $4.28$ to $2.07$, corresponding to reductions of $58.7\%$ and $51.6\%$, respectively.
Route deviations (RD) similarly decrease from $11.95$ to $8.22$, while route completion increases from $54.81$ to $56.08$.

Interestingly, SG00-FT obtains the highest infraction score ($0.88$) and particularly low vehicle-collision and red-light counts, but simultaneously exhibits substantially lower route completion ($48.22$) and the highest timeout rate ($4.31$).
This indicates that its favorable infraction statistics should be interpreted together with reduced route progress, which limits exposure to traffic interactions.
In contrast, SG10 achieves the highest route completion and lowest timeout rate among the evaluated configurations while substantially reducing several major failure categories relative to SG00-PT.
Although red-light violations remain higher than for SG00-FT, the overall results indicate a more favorable trade-off between route progress and infractions.

\begin{table*}[t!]
    \caption{\textbf{Extended infraction analysis.}
    Closed-loop LMDrive performance and infraction sub-metrics.
    SG10 is averaged across the V3 serialization configurations, while SG00-PT and SG00-FT denote the corresponding graph-free baselines.}
    \label{tab:is_analysis}
  \centering
  \footnotesize
  \begin{tabular}{c rrrrrrrrrrrrr}
    \toprule
      \multicolumn{1}{c}{Configuration} &
      \multicolumn{1}{c}{DS $\uparrow$} & \multicolumn{1}{c}{RC $\uparrow$} & \multicolumn{1}{c}{IS $\uparrow$} &
      \multicolumn{1}{c}{CP $\downarrow$} & \multicolumn{1}{c}{CV $\downarrow$} & \multicolumn{1}{c}{CL $\downarrow$} &
      \multicolumn{1}{c}{RL $\downarrow$} & \multicolumn{1}{c}{SS $\downarrow$} & \multicolumn{1}{c}{Off $\downarrow$} &
      \multicolumn{1}{c}{RD $\downarrow$} & \multicolumn{1}{c}{TO $\downarrow$} & \multicolumn{1}{c}{AB $\downarrow$} \\
    \midrule
    SG00-PT
      & 42.94 & 54.81 & 0.80 & 0.08 & 2.83 & 4.28 & 2.31 & \textbf{0.00} & 3.79 & 11.95 & 2.93 & \textbf{1.63} \\
    SG00-FT
      & 41.54 & 48.22 & \textbf{0.88} & \textbf{0.00} & \textbf{0.85} & 2.93 & \textbf{0.10} & \textbf{0.00} & 2.86 & 9.94 & 4.31 & 2.51 \\
    \midrule
    \rowcolor{gray!15} %
    SG10
      & \textbf{48.04} & \textbf{56.08} & 0.85 & 0.01 & 1.17 & \textbf{2.07} & 2.07 & 0.01 & \textbf{2.65} & \textbf{8.22} & \textbf{2.40} & 2.55 \\
    \bottomrule
  \end{tabular}

  \vspace{1.0mm}
  \raggedright
  {\fontsize{6pt}{6pt}\selectfont
  CP: Collision Pedestrians, CV: Collision Vehicles, CL: Collision Layout, RL: Red Light, SS: Stop Sign, Off: Off-Road, RD: Route Deviation, TO:~Timeout, AB: Agent Blocked.}
\end{table*}

\begin{table*}[!t]
    \caption{\textbf{Explicit relational conditioning (SG11).}
    Performance of LMDrive and BEVDriver when Actor-Only scene graphs are provided during both training and inference, evaluated across serialization formats and prompt templates V2 and V3.}
  \label{tab:sg11}
  \centering
  \footnotesize
  \begin{tabular}{c *{4}{ccc}}
    \toprule
    Serialization &
      \multicolumn{6}{c}{LMDrive} &
      \multicolumn{6}{c}{BEVDriver} \\
    \cmidrule(lr){2-7}\cmidrule(l){8-13}
    & \multicolumn{3}{c}{V2} & \multicolumn{3}{c}{V3} &
        \multicolumn{3}{c}{V2} & \multicolumn{3}{c}{V3} \\
    \cmidrule(l){2-4}\cmidrule(l){5-7}\cmidrule(l){8-10}\cmidrule(l){11-13}
    & DS $\uparrow$ & RC $\uparrow$ & IS $\uparrow$ & DS $\uparrow$ & RC $\uparrow$ & IS $\uparrow$ & DS $\uparrow$ & RC $\uparrow$ & IS $\uparrow$ & DS $\uparrow$ & RC $\uparrow$ & IS $\uparrow$ \\
    \midrule
    SG00-PT
      & 42.9 & 54.8 & 0.80 & 42.9 & 54.8 & 0.80
      & 44.7 & 49.7 & 0.90 & 44.7 & 49.7 & 0.90 \\
    SG00-FT
      & 41.5 & 48.2 & 0.88 & 41.5 & 48.2 & 0.88
      & 48.9 & 56.4 & 0.86 & 48.9 & 56.4 & 0.86 \\
    \midrule
    Text
      & 46.4 & 58.0 & 0.81 & 47.1 & 56.1 & 0.83
      & 51.7 & 61.4 & 0.84 & 52.2 & 63.5 & 0.83 \\
    JSON
      & 43.7 & 55.0 & 0.80 & \textbf{48.8} & 55.5 & 0.87
      & \textbf{53.2} & 62.8 & 0.84 & 43.8 & 51.4 & 0.88 \\
    YAML
      & 45.5 & 55.7 & 0.82 & 44.7 & 54.6 & 0.81
      & 52.7 & 61.6 & 0.85 & 52.8 & 60.4 & 0.87 \\
    \midrule
    \rowcolor{gray!15} %
    Mean
      & 45.2 & 56.2 & 0.81 & \textbf{46.9} & 55.4 & 0.84
      & \textbf{52.5} & 61.9 & 0.85 & 49.6 & 58.4 & 0.86 \\
    \bottomrule
  \end{tabular}

\end{table*}

\subsection{Explicit Relational Conditioning (SG11)}

While our primary setting uses scene graphs only during training (SG10), we additionally investigate explicit relational conditioning in which scene graphs are retained during both training and inference (SG11).
\cref{tab:sg11} reports results for LMDrive and BEVDriver using Actor-Only graphs across serialization formats and prompt templates V2 and V3.

For LMDrive, SG11 generally underperforms the corresponding SG10 configurations.
For example, the mean DS reaches $45.2$ and $46.9$ for V2 and V3, respectively, compared with $46.8$ and $48.0$ under SG10.
BEVDriver exhibits the opposite trend: SG11 reaches a mean DS of $52.5$ with V2, compared with $51.7$ for SG10, and similarly improves from $47.3$ to $49.6$ with V3.
Thus, retaining explicit scene graph context at inference can provide additional benefits, but its effectiveness depends on the underlying architecture and training configuration.

One possible explanation is the different adaptability of the language components.
LMDrive keeps its language backbone frozen and adapts primarily through the Q-Former, whereas BEVDriver additionally updates the language model using LoRA.
The latter may therefore be better able to incorporate structured relational context directly into its language representations.
Nevertheless, SG10 provides the more consistent deployment strategy across models while additionally avoiding scene graph construction and token overhead at inference.

\begin{table}[t!]
\centering
\caption{\textbf{Scene graph noise robustness.}
Comparison of SG01 and SG11 under increasing scene graph corruption, averaged across serialization formats for LMDrive using template V3.
Noise is introduced through node/edge dropout and semantic label corruption.}
\label{tab:sg_noise}
\footnotesize
\begin{tabular}{c ccc ccc}
\toprule
& \multicolumn{3}{c}{SG01} & \multicolumn{3}{c}{SG11} \\
\cmidrule(lr){2-4}\cmidrule(l){5-7}
Noise & DS & RC & IS & DS & RC & IS \\
\midrule 
None   & \textbf{44.7} & 53.8 & 0.83 & 46.9 & 55.4 & 0.84 \\
Soft   & 42.4 & 52.1 & 0.82 & 47.1 & 55.2 & 0.84 \\
Medium & 42.4 & 51.4 & 0.83 & 46.1 & 54.7 & 0.82 \\
Heavy  & 41.4 & 51.1 & 0.82 & \textbf{47.3} & 56.9 & 0.83 \\
\bottomrule
\multicolumn{7}{c}{}\\ 
\end{tabular}

\end{table}

\begin{table*}[!ht]
\caption{\textbf{Extended relational probing.}
Macro-averaged balanced accuracy (\%) for predicting scene relations from graph-free LMDrive hidden representations.
We report results for the LLM and Q-Former using a linear probe and MLP probes with hidden dimensions 128 and 256.
SG00-FT denotes graph-free fine-tuning, while JSON, Text, and YAML denote the serialization format used during SG10 training.
Random prediction yields approximately $50\%$ balanced accuracy.}
\label{tab:extended_probing}
\centering
\footnotesize
\begin{tabular}{ll cc cc cc cc}
\toprule
& &
\multicolumn{2}{c}{SG00-FT} &
\multicolumn{2}{c}{SG10-JSON} &
\multicolumn{2}{c}{SG10-Text} &
\multicolumn{2}{c}{SG10-YAML} \\
\cmidrule(lr){3-4}
\cmidrule(lr){5-6}
\cmidrule(lr){7-8}
\cmidrule(l){9-10}
Probe & Relations
& LLM & QF
& LLM & QF
& LLM & QF
& LLM & QF \\
\midrule

\multirow{3}{*}{Linear}
& All       & 77.2 & 71.7 & 77.8 & 72.3 & 77.9 & 72.1 & 78.2 & 72.0 \\
& Spatial   & 78.6 & 72.5 & 78.8 & 73.2 & 78.8 & 73.0 & 79.2 & 72.9 \\
& Proximity & 75.6 & 70.9 & 76.7 & 71.2 & 76.8 & 71.1 & 76.9 & 71.0 \\
\midrule

\multirow{3}{*}{MLP-128}
& All       & 77.7 & 73.1 & 80.2 & 72.8 & 78.8 & 72.1 & 80.3 & 73.0 \\
& Spatial   & 81.8 & 76.7 & 83.5 & 77.2 & 82.8 & 76.7 & 83.5 & 76.3 \\
& Proximity & 72.7 & 68.8 & 76.3 & 67.6 & 74.1 & 66.7 & 76.4 & 69.0 \\
\midrule

\multirow{3}{*}{MLP-256}
& All       & 77.1 & 73.1 & 79.3 & 74.5 & 79.3 & 72.8 & 79.8 & 73.4 \\
& Spatial   & 81.6 & 77.1 & 83.7 & 78.4 & 82.7 & 76.9 & 83.0 & 76.8 \\
& Proximity & 71.6 & 68.2 & 74.0 & 69.8 & 75.3 & 67.9 & 76.1 & 69.2 \\
\bottomrule
\end{tabular}
\end{table*}

\begin{table*}[t!]
\caption{\textbf{Relation-wise probing with MLP-256.}
Balanced accuracy (\%) for individual scene relations using a representative nonlinear probe with hidden dimension 256.
Results are reported for graph-free LLM and Q-Former representations.
SG10 denotes the serialization format used only during training.}
\label{tab:relationwise_probing}
\centering
\footnotesize
\begin{tabular}{l cc cc cc cc}
\toprule
&
\multicolumn{2}{c}{SG00-FT} &
\multicolumn{2}{c}{SG10-JSON} &
\multicolumn{2}{c}{SG10-Text} &
\multicolumn{2}{c}{SG10-YAML} \\
\cmidrule(lr){2-3}
\cmidrule(lr){4-5}
\cmidrule(lr){6-7}
\cmidrule(l){8-9}
Relation
& LLM & QF
& LLM & QF
& LLM & QF
& LLM & QF \\
\midrule
atDRearOf    & 86.3 & 81.4 & 87.9 & 81.6 & 87.8 & 81.3 & 87.8 & 80.9 \\
atSRearOf    & 72.2 & 68.0 & 76.5 & 71.0 & 74.7 & 67.9 & 74.3 & 66.1 \\
inDFrontOf   & 87.3 & 82.4 & 87.4 & 82.4 & 86.8 & 80.2 & 87.3 & 82.0 \\
inSFrontOf   & 77.8 & 75.5 & 79.2 & 76.0 & 79.6 & 74.2 & 79.9 & 75.3 \\
toLeftOf     & 82.7 & 76.7 & 85.9 & 78.9 & 83.3 & 77.6 & 83.4 & 77.7 \\
toRightOf    & 83.2 & 78.7 & 85.1 & 80.5 & 83.9 & 79.9 & 85.0 & 78.7 \\
near         & 79.6 & 74.3 & 81.5 & 76.2 & 82.6 & 76.1 & 82.8 & 74.3 \\
near\_coll   & 54.9 & 58.8 & 80.1 & 62.6 & 61.9 & 57.6 & 66.3 & 62.4 \\
super\_near  & 65.3 & 59.9 & 65.3 & 58.5 & 70.4 & 57.9 & 68.0 & 61.0 \\
very\_near   & 77.7 & 69.7 & 79.9 & 73.2 & 79.1 & 70.5 & 81.0 & 69.7 \\
visible      & 80.6 & 78.2 & 83.3 & 78.5 & 82.5 & 77.5 & 82.2 & 78.9 \\
\midrule
All          & 77.1 & 73.1 & 79.3 & 74.5 & 79.3 & 72.8 & 79.8 & 73.4 \\
Spatial      & 81.6 & 77.1 & 83.7 & 78.4 & 82.7 & 76.9 & 83.0 & 76.8 \\
Proximity    & 71.6 & 68.2 & 74.0 & 69.8 & 75.3 & 67.9 & 76.1 & 69.2 \\
\bottomrule
\end{tabular}
\end{table*}

\subsection{Scene Graph Noise Robustness}

Finally, we assess sensitivity to imperfect scene graph inputs by corrupting the graphs provided at inference.
We simulate perception and automatic-labeling errors through increasing node and edge dropout ($20\%$--$60\%$) and semantic label corruption ($10\%$--$30\%$).
We compare zero-shot graph injection (SG01) with models explicitly trained and evaluated with scene graphs (SG11), averaging results across serialization formats for LMDrive with template V3.

As shown in \cref{tab:sg_noise}, SG01 gradually degrades as corruption increases, with DS decreasing from $44.7$ without noise to $41.4$ under heavy corruption.
In contrast, SG11 remains comparatively stable across all tested noise levels, with DS ranging from $46.1$ to $47.3$.
This suggests that prior exposure to structured relational inputs during training makes the model less sensitive to moderate perturbations of the graph at inference.

The robustness may additionally benefit from the discrete, coarse-grained relation vocabulary used by our scene graphs, which abstracts continuous geometric measurements into semantic relations.
However, the present experiment does not isolate the contribution of this discretization from the effect of scene graph-conditioned training.
Importantly, our primary SG10 setting is unaffected by test-time scene graph corruption altogether, since no graph is required during inference.

\subsection{Extended Relational Probing}

We provide additional probing results to characterize where relational information becomes accessible under graph-free inference.
\cref{tab:extended_probing} extends the main-paper analysis to both the LLM and Q-Former representations and separates spatial and proximity relations.
Across all three probe architectures, SG10 consistently increases the decodability of relational information from the LLM hidden states.
For example, macro balanced accuracy increases from $77.2\%$ to $77.8$--$78.2\%$ with the linear probe, from $77.7\%$ to $78.8$--$80.3\%$ with MLP-128, and from $77.1\%$ to $79.3$--$79.8\%$ with MLP-256.
In contrast, changes in the Q-Former representations are smaller and less consistent across serialization formats and probe capacities.

For a relation-wise analysis, \cref{tab:relationwise_probing} reports results using the MLP-256 probe.
The improvement in LLM representations is observed across many spatial and proximity relations, with particularly pronounced gains for proximity-related interactions.
The proximity macro average increases from $71.6\%$ for SG00-FT to $74.0\%$, $75.3\%$, and $76.1\%$ for JSON, Text, and YAML SG10 training, respectively.
The largest individual change occurs for \texttt{near\_coll}, which increases from $54.9\%$ to $80.1\%$ with JSON supervision.
These results indicate that the aggregate probing gains are not solely driven by generic scene information, but are reflected in increased accessibility of specific interaction relations.

\begin{table*}[!ht]
  \caption{\textbf{Extended baseline performance.}
Track-wise performance of LMDrive and BEVDriver without scene graphs.
SG00-PT denotes the officially released checkpoints, while SG00-FT further fine-tunes the same checkpoints on the extended LMDrive training split used for our scene graph-conditioned experiments.}
  \label{tab:langauto_lengths}
  \centering
  \fontsize{7pt}{7pt}\selectfont
  \begin{tabular}{cc*9{c}}
    \toprule
    Method & Config &
    \multicolumn{3}{c}{Tiny} &
    \multicolumn{3}{c}{Short} &
    \multicolumn{3}{c}{Long} \\
    \cmidrule(l){3-5}\cmidrule(l){6-8}\cmidrule(l){9-11} &
     & \multicolumn{1}{c}{DS $\uparrow$} & \multicolumn{1}{c}{RC $\uparrow$} & \multicolumn{1}{c}{IS $\uparrow$}
     & \multicolumn{1}{c}{DS $\uparrow$} & \multicolumn{1}{c}{RC $\uparrow$} & \multicolumn{1}{c}{IS $\uparrow$}
     & \multicolumn{1}{c}{DS $\uparrow$} & \multicolumn{1}{c}{RC $\uparrow$} & \multicolumn{1}{c}{IS $\uparrow$} \\
    \midrule
    \multirow{2}{*}{LMDrive} & SG00-PT   & 60.72 & 70.98 & 0.82 & 41.34 & 56.98 & 0.79 & 26.80 & 36.47 & 0.77 \\
     & SG00-FT   & 57.98 & 65.83 & 0.88 & 41.69 & 50.25 & 0.86 & 24.98 & 28.58 & 0.90 \\
    \midrule
    \multirow{2}{*}{BEVDriver} & SG00-PT & 63.15 & 68.31 & 0.92 & 42.22 & 46.00 & 0.92 & 28.74 & 34.79 & 0.86 \\
     & SG00-FT & 55.80 & 62.22 & 0.88 & 62.27 & 69.35 & 0.89 & 28.75 & 37.53 & 0.82 \\
    \bottomrule
  \end{tabular}
\end{table*}

\begin{table*}[!ht]
  \caption{\textbf{Extended zero-shot scene graph injection (SG01).}
Track-wise LMDrive performance across serialization formats, graph abstraction levels, and prompt templates under test-time-only scene graph conditioning.}
  \label{tab:lmdrive_combined}
  \centering
  \fontsize{7pt}{7pt}\selectfont
  \begin{tabular}{c c c *{3}{ccc}}
    \toprule
    \multicolumn{1}{c}{Serialization} &
    \multicolumn{1}{c}{Graph Abstraction} &
    \multicolumn{1}{c}{Template} &
    \multicolumn{3}{c}{Tiny} &
    \multicolumn{3}{c}{Short} &
    \multicolumn{3}{c}{Long} \\
    \cmidrule(l){4-6}\cmidrule(l){7-9}\cmidrule(l){10-12}
    & & &
    DS $\uparrow$ & RC $\uparrow$ & IS $\uparrow$ &
    DS $\uparrow$ & RC $\uparrow$ & IS $\uparrow$ &
    DS $\uparrow$ & RC $\uparrow$ & IS $\uparrow$ \\
    \midrule

    \multirow{9}{*}{Text}
      & \multirow{3}{*}{Full} & V1
        & 49.6 & 56.0 & 0.87 & 44.7 & 54.1 & 0.85 & 22.5 & 28.5 & 0.81 \\
      &  & V2
        & 59.3 & 64.4 & 0.89 & 42.1 & 52.8 & 0.83 & 23.9 & 31.0 & 0.77 \\
      &  & V3
        & 58.5 & 66.5 & 0.86 & 40.0 & 49.5 & 0.84 & 25.3 & 33.1 & 0.79 \\

      \cmidrule{2-12}

      & \multirow{3}{*}{Road-Level} & V1
        & 54.2 & 60.8 & 0.86 & 43.8 & 52.0 & 0.86 & 23.8 & 28.3 & 0.84 \\
      &  & V2
        & 58.2 & 64.1 & 0.89 & 43.5 & 54.1 & 0.82 & 22.9 & 30.0 & 0.77 \\
      &  & V3
        & 58.3 & 64.1 & 0.87 & 41.1 & 56.8 & 0.80 & 22.1 & 29.3 & 0.75 \\

      \cmidrule{2-12}

      & \multirow{3}{*}{Actor-Only} & V1
        & 50.9 & 60.1 & 0.85 & 42.9 & 52.4 & 0.86 & 26.3 & 33.7 & 0.83 \\
      &  & V2
        & 64.7 & 71.3 & 0.89 & 44.8 & 53.1 & 0.87 & 24.5 & 34.5 & 0.77 \\
      &  & V3
        & 59.8 & 65.7 & 0.89 & 45.3 & 56.4 & 0.83 & 24.2 & 31.4 & 0.80 \\
    \midrule

    \multirow{9}{*}{JSON}

      & \multirow{3}{*}{Full} & V1
        & 45.8 & 51.4 & 0.87 & 34.6 & 46.3 & 0.81 & 23.4 & 28.8 & 0.83 \\
      &  & V2
        & 61.5 & 66.9 & 0.90 & 39.2 & 49.7 & 0.84 & 23.7 & 29.4 & 0.79 \\
      &  & V3
        & 62.2 & 69.3 & 0.88 & 42.0 & 55.0 & 0.80 & 29.7 & 39.6 & 0.80 \\

      \cmidrule{2-12}

      & \multirow{3}{*}{Road-Level} & V1
        & 44.2 & 49.3 & 0.88 & 38.1 & 54.0 & 0.78 & 21.0 & 26.8 & 0.82 \\
      &  & V2
        & 59.4 & 66.6 & 0.86 & 41.7 & 56.8 & 0.77 & 24.3 & 30.1 & 0.78 \\
      &  & V3
        & 62.7 & 71.3 & 0.86 & 49.5 & 59.8 & 0.85 & 27.7 & 37.9 & 0.75 \\

      \cmidrule{2-12}

      & \multirow{3}{*}{Actor-Only} & V1
        & 55.6 & 60.8 & 0.89 & 41.5 & 51.0 & 0.86 & 23.9 & 30.1 & 0.81 \\
      &  & V2
        & 60.2 & 64.7 & 0.91 & 42.7 & 58.1 & 0.78 & 25.0 & 34.1 & 0.76 \\
      &  & V3
        & 58.7 & 71.0 & 0.82 & 48.0 & 58.1 & 0.86 & 27.5 & 37.5 & 0.77 \\
    \midrule

    \multirow{9}{*}{YAML}

      & \multirow{3}{*}{Full} & V1
        & 49.4 & 54.2 & 0.87 & 47.0 & 54.1 & 0.88 & 21.7 & 28.1 & 0.81 \\
      &  & V2
        & 60.5 & 66.0 & 0.90 & 41.0 & 54.7 & 0.80 & 23.9 & 30.7 & 0.76 \\
      &  & V3
        & 56.7 & 62.7 & 0.87 & 43.5 & 58.7 & 0.79 & 25.2 & 34.2 & 0.74 \\

      \cmidrule{2-12}

      & \multirow{3}{*}{Road-Level} & V1
        & 52.3 & 58.0 & 0.87 & 43.1 & 53.8 & 0.86 & 22.3 & 28.0 & 0.82 \\
      &  & V2
        & 59.3 & 64.8 & 0.90 & 41.4 & 58.3 & 0.77 & 24.8 & 31.6 & 0.79 \\
      &  & V3
        & 57.9 & 63.8 & 0.88 & 43.5 & 54.4 & 0.84 & 24.9 & 33.7 & 0.77 \\

      \cmidrule{2-12}

      & \multirow{3}{*}{Actor-Only} & V1
        & 60.9 & 69.3 & 0.86 & 43.2 & 56.2 & 0.82 & 24.3 & 30.6 & 0.82 \\
      &  & V2
        & 60.6 & 69.3 & 0.88 & 44.8 & 54.0 & 0.85 & 25.2 & 36.1 & 0.72 \\
      &  & V3
        & 62.1 & 67.5 & 0.91 & 46.7 & 58.1 & 0.84 & 25.3 & 32.6 & 0.78 \\
    \bottomrule
  \end{tabular}

    \vspace{-0.5em}
  
\end{table*}

\section{Detailed Quantitative Results}

This section provides extended benchmark-wise results for the configurations summarized in the main paper.
While the main paper reports aggregate performance across tracks, we provide the corresponding \textit{Tiny}, \textit{Short}, and \textit{Long} breakdowns for completeness.

\subsection{Extended Baseline Performance}

\cref{tab:langauto_lengths} reports track-wise performance for LMDrive and BEVDriver without scene graph input, comparing the official pretrained checkpoints (SG00-PT) with the corresponding graph-free fine-tuned models (SG00-FT).
These results provide the \textit{Tiny}, \textit{Short}, and \textit{Long} breakdowns underlying the aggregate baselines reported in the main paper.

\begin{table*}[!ht]
\caption{\textbf{Extended scene graph-conditioned training results for LMDrive.}
Track-wise performance of SG10 and SG11 using Actor-Only graphs across serialization formats and prompt templates V2 and V3.}
  \label{tab:langauto_lmdrive_templates}
  \centering
  \fontsize{7pt}{7pt}\selectfont
  \begin{tabular}{c c c *{3}{ccc}}
    \toprule
    \multicolumn{1}{c}{Serialization} &
    \multicolumn{1}{c}{Template} &
    \multicolumn{1}{c}{Config} &
    \multicolumn{3}{c}{Tiny} &
    \multicolumn{3}{c}{Short} &
    \multicolumn{3}{c}{Long} \\
    \cmidrule(l){4-6}\cmidrule(l){7-9}\cmidrule(l){10-12}
    & & &
    DS $\uparrow$ & RC $\uparrow$ & IS $\uparrow$ &
    DS $\uparrow$ & RC $\uparrow$ & IS $\uparrow$ &
    DS $\uparrow$ & RC $\uparrow$ & IS $\uparrow$ \\
    \midrule

    \multirow{4}{*}{Text}
      & \multirow{2}{*}{V2} & SG11
        & 66.6 & 76.1 & 0.88 & 45.5 & 60.9 & 0.77 & 27.2 & 36.9 & 0.76 \\
      &  & SG10
        & 66.8 & 75.2 & 0.88 & 46.4 & 57.8 & 0.82 & 27.3 & 35.4 & 0.81 \\ \cmidrule{2-12}
      & \multirow{2}{*}{V3} & SG11
        & 52.8 & 62.0 & 0.85 & 56.3 & 65.8 & 0.87 & 32.0 & 40.5 & 0.78 \\
      &  & SG10
        & 72.6 & 78.7 & 0.91 & 52.9 & 60.6 & 0.87 & 29.9 & 39.0 & 0.81 \\
    \midrule

    \multirow{4}{*}{JSON}
      & \multirow{2}{*}{V2} & SG11
        & 55.2 & 67.2 & 0.82 & 45.9 & 57.7 & 0.81 & 30.1 & 40.2 & 0.77 \\
      &  & SG10
        & 57.7 & 67.6 & 0.86 & 53.3 & 59.7 & 0.88 & 26.1 & 33.2 & 0.81 \\ \cmidrule{2-12}
      & \multirow{2}{*}{V3} & SG11
        & 64.9 & 68.6 & 0.94 & 51.1 & 59.1 & 0.84 & 30.5 & 39.0 & 0.82 \\
      &  & SG10
        & 53.5 & 60.2 & 0.89 & 53.9 & 62.9 & 0.85 & 29.2 & 38.8 & 0.81 \\
    \midrule

    \multirow{4}{*}{YAML}
      & \multirow{2}{*}{V2} & SG11
        & 61.3 & 67.1 & 0.90 & 47.5 & 61.2 & 0.79 & 27.8 & 38.8 & 0.75 \\
      &  & SG10
        & 63.8 & 72.1 & 0.88 & 49.1 & 58.1 & 0.83 & 30.9 & 42.1 & 0.79 \\ \cmidrule{2-12}
      & \multirow{2}{*}{V3} & SG11
        & 61.9 & 71.4 & 0.86 & 43.7 & 53.7 & 0.79 & 28.6 & 38.8 & 0.77 \\
      &  & SG10
        & 63.7 & 71.0 & 0.89 & 50.9 & 56.8 & 0.87 & 25.8 & 36.8 & 0.75 \\
    \bottomrule
  \end{tabular}
\end{table*}

\begin{table*}[!ht]
\caption{\textbf{Extended scene graph-conditioned training results for BEVDriver.}
Track-wise performance of SG10 and SG11 using Actor-Only graphs across serialization formats and prompt templates V2 and V3.}
  \label{tab:langauto_bevdriver_templates}
  \centering
  \fontsize{7pt}{7pt}\selectfont
  \begin{tabular}{c c c *{3}{ccc}}
    \toprule
    \multicolumn{1}{c}{Serialization} &
    \multicolumn{1}{c}{Template} &
    \multicolumn{1}{c}{Config} &
    \multicolumn{3}{c}{Tiny} &
    \multicolumn{3}{c}{Short} &
    \multicolumn{3}{c}{Long} \\
    \cmidrule(l){4-6}\cmidrule(l){7-9}\cmidrule(l){10-12}
    & & &
    DS $\uparrow$ & RC $\uparrow$ & IS $\uparrow$ &
    DS $\uparrow$ & RC $\uparrow$ & IS $\uparrow$ &
    DS $\uparrow$ & RC $\uparrow$ & IS $\uparrow$ \\
    \midrule

    \multirow{4}{*}{Text}
      & \multirow{2}{*}{V2} & SG11
        & 60.7 & 69.0 & 0.88 & 59.2 & 71.5 & 0.83 & 35.2 & 43.7 & 0.81 \\
      &  & SG10
        & 66.9 & 78.5 & 0.85 & 50.0 & 54.5 & 0.93 & 38.2 & 46.5 & 0.83 \\ \cmidrule{2-12}
      & \multirow{2}{*}{V3} & SG11
        & 61.5 & 71.1 & 0.88 & 60.8 & 70.4 & 0.88 & 34.2 & 49.0 & 0.73 \\
      &  & SG10
        & 61.3 & 67.8 & 0.90 & 59.5 & 67.8 & 0.89 & 33.7 & 44.4 & 0.78 \\
    \midrule

    \multirow{4}{*}{JSON}
      & \multirow{2}{*}{V2} & SG11
        & 60.9 & 69.4 & 0.88 & 63.9 & 73.3 & 0.88 & 34.7 & 45.7 & 0.76 \\
      &  & SG10
        & 52.0 & 57.1 & 0.89 & 54.9 & 58.5 & 0.94 & 35.1 & 40.7 & 0.85 \\ \cmidrule{2-12}
      & \multirow{2}{*}{V3} & SG11
        & 49.3 & 56.3 & 0.89 & 47.2 & 55.0 & 0.90 & 34.9 & 43.0 & 0.83 \\
      &  & SG10
        & 49.2 & 56.9 & 0.90 & 48.4 & 55.9 & 0.89 & 29.1 & 39.3 & 0.77 \\
    \midrule

    \multirow{4}{*}{YAML}
      & \multirow{2}{*}{V2} & SG11
        & 58.0 & 64.0 & 0.91 & 64.1 & 72.8 & 0.87 & 36.1 & 48.1 & 0.78 \\
      &  & SG10
        & 58.8 & 69.2 & 0.86 & 71.7 & 76.2 & 0.94 & 37.7 & 46.8 & 0.81 \\ \cmidrule{2-12}
      & \multirow{2}{*}{V3} & SG11
        & 58.4 & 67.8 & 0.86 & 58.5 & 63.6 & 0.90 & 41.7 & 49.7 & 0.84 \\
      &  & SG10
        & 58.8 & 66.4 & 0.86 & 55.3 & 62.9 & 0.87 & 30.5 & 37.7 & 0.81 \\
    \bottomrule
  \end{tabular}
\end{table*}

\begin{table*}[t!]
\caption{\textbf{Extended SpaceDrive results on nuScenes.}
Open-loop planning performance at 1\,s, 2\,s, and 3\,s prediction horizons.
SG00-FT and SG10 are fine-tuned for two additional epochs from the same official checkpoint under matched optimization settings, differing only in scene graph supervision during training.}
\label{tab:spacedrive_extended}
\centering
\fontsize{7pt}{7pt}\selectfont

\begin{tabular}{l
                cccc
                cccc
                cccc
                c}
\toprule
& \multicolumn{4}{c}{L2 $\downarrow$}
& \multicolumn{4}{c}{Collision $\downarrow$}
& \multicolumn{4}{c}{Intersection $\downarrow$}
& Overall $\downarrow$ \\
\cmidrule(lr){2-5}
\cmidrule(lr){6-9}
\cmidrule(lr){10-13}
\cmidrule(l){14-14}

Method
& 1s & 2s & 3s & Avg.
& 1s & 2s & 3s & Avg.
& 1s & 2s & 3s & Avg.
& Avg. \\
\midrule

SG00-PT
& 0.152 & 0.294 & 0.516 & 0.321
& 0.039 & 0.176 & 0.469 & 0.228
& 0.234 & 0.820 & 2.891 & 1.315
& 0.621 \\

\midrule

SG00-FT
& 0.160 & 0.305 & 0.533 & 0.332
& \textbf{0.000} & \textbf{0.117} & 0.469 & \textbf{0.195}
& 0.215 & 0.723 & 3.399 & 1.446
& 0.658 \\

\rowcolor{gray!15}
SG10, Text
& \textbf{0.153} & \textbf{0.297} & \textbf{0.525} & \textbf{0.325}
& 0.059 & 0.176 & \textbf{0.447} & 0.227
& \textbf{0.195} & \textbf{0.684} & \textbf{2.696} & \textbf{1.192}
& \textbf{0.581} \\

\bottomrule
\end{tabular}
\end{table*}

\subsection{Extended Zero-Shot Scene Graph Injection (SG01)}

\cref{tab:lmdrive_combined} provides the complete track-wise SG01 results for LMDrive across serialization formats, abstraction levels, and prompt templates.
While the main paper reports performance averaged across LangAuto tracks, this breakdown shows the corresponding results on \textit{Tiny}, \textit{Short}, and \textit{Long} individually.

\subsection{Extended Scene Graph-Conditioned Training Results}
\cref{tab:langauto_lmdrive_templates,tab:langauto_bevdriver_templates} provide the complete track-wise results for LMDrive and BEVDriver under SG10 and SG11.
All configurations use Actor-Only graphs and evaluate Text, JSON, and YAML serialization with prompt templates V2 and V3.
While the main paper reports aggregate performance across LangAuto tracks, these tables provide the corresponding \textit{Tiny}, \textit{Short}, and \textit{Long} breakdowns.

\subsection{Extended SpaceDrive Results}

\cref{tab:spacedrive_extended} provides the horizon-wise breakdown corresponding to the SpaceDrive experiment reported in the main paper.
Starting from the same official checkpoint, SG00-FT and SG10 are fine-tuned for two additional epochs under identical optimization settings, differing only in the presence of scene graph supervision during training.
SG10 improves L2 error at all prediction horizons and substantially reduces the intersection metric, while the collision results are mixed across horizons.

\section{Qualitative Results}

Alongside this supplementary material, we include a video providing a brief overview of our method and a comparison of the different scene graph abstraction levels (Full, Road-Level, and Actor-Only) for a representative driving scenario. 
We then showcase qualitative closed-loop results for LMDrive, BEVDriver, and SimLingo.
The video highlights specific scenarios where the original pretrained baselines fail due to collisions or incorrect instruction following, and demonstrates how our scene graph-conditioned models successfully navigate these challenges through internalized relational reasoning.

\end{document}